\documentclass[10pt,twocolumn,letterpaper]{article}

\usepackage[pagenumbers]{cvpr}   

\makeatletter
\@namedef{ver@everyshi.sty}{}
\makeatother

\usepackage{graphicx}
\usepackage{amsmath}
\usepackage{amssymb}
\usepackage{booktabs}
\usepackage{comment}
\usepackage[notransparent]{svg}
\usepackage{bbding}
\usepackage{makecell}
\usepackage{multirow}
\usepackage{tabularx}
\usepackage[page]{appendix} 

\usepackage{titletoc}

\usepackage[pagebackref,breaklinks,colorlinks]{hyperref}

\usepackage[capitalize]{cleveref}
\crefname{section}{Sec.}{Secs.}
\Crefname{section}{Section}{Sections}
\Crefname{table}{Table}{Tables}
\crefname{table}{Tab.}{Tabs.}

\usepackage[labelsep=period]{caption}
\renewcommand{\paragraph}[1]{\vspace{0.2em}\noindent \textbf{#1 \hspace{0.2em}}}

\definecolor{MyDarkRed}{rgb}{0.66, 0.16, 0.16}
\definecolor{MyDarkBlue}{rgb}{0.16, 0.16, 0.66}
\usepackage{lipsum} 

\definecolor{babyblue}{rgb}{0.54, 0.81, 0.94}

\definecolor{greensheen}{RGB}{128, 173, 160}

\usepackage[marginal]{footmisc}
\renewcommand{\thefootnote}{\fnsymbol{footnote}}
\usepackage{url}
\usepackage{colortbl}

\newcommand\blfootnote[1]{%
  \begingroup
  \renewcommand\thefootnote{}\footnote{#1}%
  \addtocounter{footnote}{-1}%
  \endgroup
}

\def\cvprPaperID{2010} 
\def\confName{CVPR}
\def\confYear{2023}

\begin{document}

\title{MVImgNet: A Large-scale Dataset of Multi-view Images}

\author{
Xianggang Yu$^{*}$ \qquad Mutian Xu$^{*,\dag}$ \qquad Yidan Zhang$^{*}$ \qquad Haolin Liu$^{*}$ \qquad Chongjie Ye$^{*}$ \\ Yushuang Wu \qquad Zizheng Yan \qquad Chenming Zhu \qquad Zhangyang Xiong \qquad Tianyou Liang \\ Guanying Chen \qquad Shuguang Cui \qquad Xiaoguang Han$^{\ddag}$ \vspace{5pt}\\
\small{$^{*}$equal technical contribution} \qquad \small{$^{\dag}$part of project lead} \qquad \small{$^{\ddag}$corresponding author} \vspace{5pt}\\
{SSE, CUHKSZ} \qquad {FNii, CUHKSZ} \vspace{5pt}\\
\small{\href{https://gaplab.cuhk.edu.cn/projects/MVImgNet/}{gaplab.cuhk.edu.cn/projects/MVImgNet}}
}

\twocolumn[{%
\renewcommand\twocolumn[1][]{#1}%
\maketitle
\begin{center}
    \centering
    \captionsetup{type=figure}
    \includegraphics[width=1.0\textwidth]{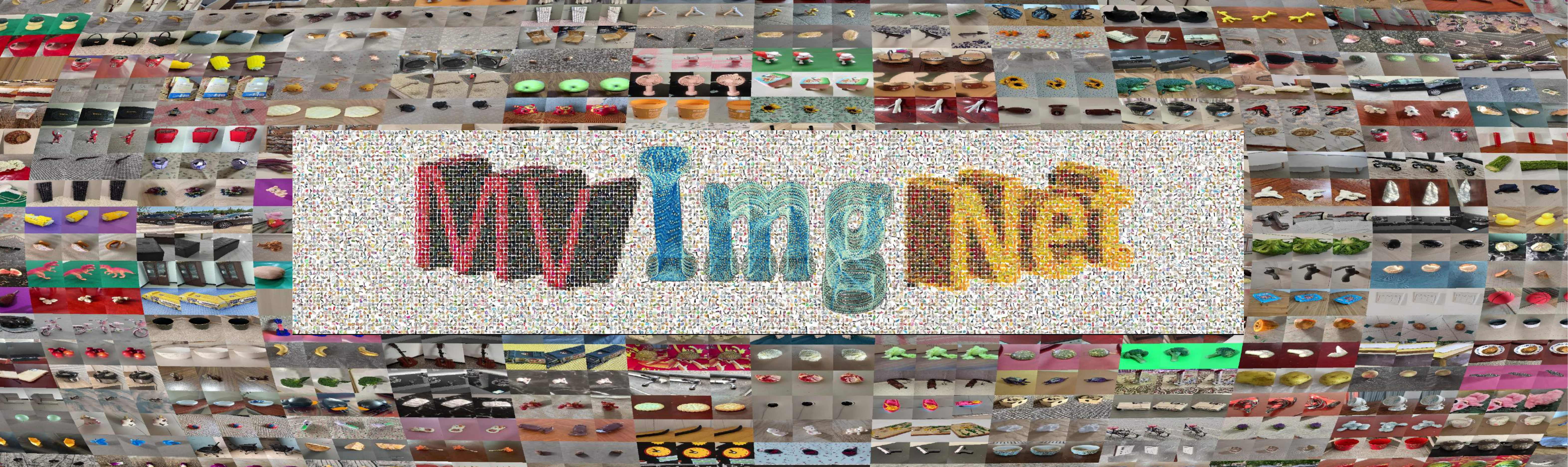}
    \captionof{figure}{It is time to embrace \textbf{MVImgNet}!
    We introduce MVImgnet, a large-scale dataset of \textit{multi-view images}, which is efficiently collected by shooting videos of real-world objects. 
    It enjoys 3D-aware signals from multi-view consistency, being a soft bridge between 2D and 3D vision.
    Through dense reconstruction on MVImgNet, we also present a large-scale real-world 3D object \textit{point cloud} dataset -- MVPNet.
    \textbf{Exterior}: Examples of various multi-view images in MVImgNet (see Fig.~\ref{fig:pano} more intuitively).
    \textbf{Interior}: Instances of colorful point clouds from MVPNet (see Fig.~\ref{fig:mvppano} more clearly) are assembled into a stereo sign of `MVImgNet'.
    \label{fig:intro}}
\end{center}%
}]
\blfootnote{Author contributions \hyperref[sec:Contributions]{listed at end of the paper}.}

\begin{abstract}
    Being data-driven is one of the most iconic properties of deep learning algorithms.
    The birth of ImageNet \cite{imagenet} drives a remarkable trend of `learning from large-scale data' in computer vision.
    Pretraining on ImageNet to obtain rich universal representations has been manifested to benefit various 2D visual tasks, and becomes a standard in 2D vision.
    However, due to the laborious collection of real-world 3D data, there is yet no generic dataset serving as a counterpart of ImageNet in 3D vision, thus how such a dataset can impact the 3D community is unraveled.
    To remedy this defect, we introduce \textbf{MVImgNet}, a large-scale dataset of multi-view images, which is highly convenient to gain by shooting videos of real-world objects in human daily life.
    It contains \textbf{6.5 million} frames from \textbf{219,188} videos crossing objects from \textbf{238} classes, with rich annotations of object masks, camera parameters, and point clouds.
    The multi-view attribute endows our dataset with 3D-aware signals, making it a soft bridge between 2D and 3D vision.
    
    We conduct pilot studies for probing the potential of MVImgNet on a variety of 3D and 2D visual tasks, including radiance field reconstruction, multi-view stereo, and view-consistent image understanding, where MVImgNet demonstrates promising performance, remaining lots of possibilities for future explorations.
    
    Besides, via dense reconstruction on MVImgNet, a 3D object point cloud dataset is derived, called \textbf{MVPNet}, covering \textbf{87,200} samples from \textbf{150} categories, with the class label on each point cloud. Experiments show that MVPNet can benefit the real-world 3D object classification while posing new challenges to point cloud understanding.

    MVImgNet and MVPNet will be public, hoping to inspire the broader vision community.
    
    \vspace{-0.3cm}
\end{abstract}

\section{Introduction}
\label{sec:intro}
Being data-driven, also known as data-hungry, is one of the most important attributes of deep learning algorithms.
By training on large-scale datasets,
deep neural networks are able to extract rich representations.
In the past few years, the computer vision community has witnessed the bloom of such \textit{`learning from data'} regime \cite{krizhevsky2012imagenet,he2015delving,he2016deep}, after the birth of ImageNet \cite{imagenet} -- the pioneer of large-scale real-world image datasets. 
Notably, pretraining on ImageNet is well-proven to boost the model performance when transferring the pretrained weights into not only high-level \cite{rethinkpre,long2015fully,maskrcnn,girshick2015fast,liu2021Swin} but also low-level visual tasks \cite{chen2021pre,liang2021swinir}, and becomes a \textit{de-facto} standard in 2D. Recently, various 3D datasets \cite{modelnet,shapenet,s3dis,scannet,toscene,kitti,nuscenes} are produced to facilitate 3D visual applications.


However, due to the non-trivial scanning and laborious labeling of real-world 3D data (commonly organized in point clouds or meshes), existing 3D datasets are either synthetic or their scales are not comparable with ImageNet \cite{imagenet}.
Consequently, unlike in 2D vision where models are usually pretrained on ImageNet to gain universal representation or commonsense knowledge, most of the current methods in 3D area are directly trained and evaluated on particular datasets for solving specific 3D visual tasks (\eg, NeRF dataset~\cite{nerf} and ShapeNet~\cite{shapenet} for novel view synthesis, ModelNet \cite{modelnet} and ScanObjectNN \cite{scanobjectnn} for object classification, KITTI\cite{kitti} and ScanNet\cite{scannet} for scene understanding).
Here, two crucial and successive issues can be induced:
\textbf{(1)} \textit{There is still no \textbf{generic dataset} in 3D vision, as a counterpart of ImageNet in 2D.}
\textbf{(2)} \textit{What \textbf{benefit} such a dataset can endow to 3D community is yet unknown}.
In this paper, we focus on investigating these two problems and set two corresponding targets: Build the primary dataset, then explore its effect through experiments.

\textbf{\textit{Milestone 1 --} Dataset:}

For a clearer picture of the first goal, we start by carefully revisiting existing 3D datasets as well as ImageNet \cite{imagenet}.
\textbf{i)}~3D synthetic datasets \cite{modelnet,shapenet} provide rich 3D CAD models. However, they lack \textit{real-world} cues (\eg, context, occlusions, noises), which are indispensable for model robustness in practical applications. 
ScanObjectNN \cite{scanobjectnn} extracts real-world 3D objects from indoor scene data, but is limited in scale.
For 3D scene-level dataset \cite{sun-rgbd,scannet,s3dis,matterport3d,semantic3d,kitti}, their scales are still constrained by the laborious scanning and labeling (\eg, millions of points per scene). Additionally, they contain specific inner-domain knowledge such as a particularly intricate indoor room or outdoor driving configurations, making it hard for general transfer learning.
\textbf{ii)}~Although ImageNet \cite{imagenet} contains the most comprehensive real-world objects, it only describes a 2D world that misses \textit{3D-aware} signals. Since humans live in a 3D world, 3D consciousness is vitally important for realizing human-like intelligence and solving real-life visual problems. 

Based on the above review, our dataset is created from a new insight -- \textbf{multi-view images}, as a soft bridge between 2D and 3D. It lies several benefits to remedying the aforementioned defects.
Such data can be \textit{easily gained} in \textit{considerable sizes} via shooting an object around different views on common mobile devices with cameras (\eg, smartphones), which can be collected by crowd-sourcing in \textit{real world}.
Moreover, the multi-view constraint can bring natural 3D visual signals (later experiments show that this not only benefits 3D tasks but also 2D image understanding).
To this end, we build \textbf{MVImgNet}, containing \textbf{6.5} million frames from \textbf{219,188} videos crossing real-life objects from \textbf{238} classes, with rich annotations of object masks, camera parameters, and point clouds.
You may take a glance at our MVImgNet from Fig.~\ref{fig:intro}.

\textbf{\textit{Milestone 2 --} Experimental Exploration:}

Now facing the second goal of this paper, we attempt to probe the power of our dataset by conducting some pilot experiments.
Leveraging the multi-view nature of the data, we start by focusing on the view-based 3D reconstruction task and demonstrate that pretraining on MVImgNet can not only benefit the \textit{generalization ability of NeRF} (Sec.~\ref{sub:nerf}), but also \textit{data-efficient} \textit{multi-view stereo} (Sec.~\ref{sub:mvs}).
Moreover, for image understanding, although humans can easily recognize one object from different viewpoints, deep learning models can hardly do that robustly \cite{dong2022viewfool}.
Considering MVImgNet provides numerous images of a particular object from different viewpoints, we verify that MVImgNet-pretrained models are endowed with decent \textit{view consistency} in general \textit{image classification} (supervised learning in Sec.~\ref{sub:img_cls}, self-supervised contrastive learning in Sec.~\ref{sub:contras}) and \textit{salient object detection} (Sec.~\ref{sub:sod}).

\textbf{\textit{Bonus --} A New 3D Point Cloud Dataset -- MVPNet:}

Through dense reconstruction on MVImgNet, a new 3D object \textbf{p}oint cloud dataset is derived, called \textbf{MVPNet}, which contains \textbf{87,200} point clouds with 150 categories, with the class label on each point cloud (see \cref{fig:mvppano}). 
Experiments show that MVPNet not only benefits the real-world 3D object classification task but also poses new challenges and prospects to point cloud understanding (Sec.~\ref{sec:3d_understand}).

MVImgNet and MVPNet will be public, hoping to inspire the broader vision community.

\section{Related Work}
\label{sec:related_works}




\paragraph{Single-view image datasets.} 
The MNIST database~\cite{mnist} is one of the most pioneering datasets,
composed of 70k monochrome images of handwritten digits. 
The CIFAR10 and CIFAR100 datasets proposed by \cite{cifar}, respectively collect 60k tiny color images (32$\times$32) of various common objects or animals in 10 and 100 classes. 
ImageNet~\cite{imagenet} is presented with a large scale, high accuracy, large diversity, and hierarchical structure, which provides opportunities for training deep neural networks.
In detection and segmentation tasks, MSCOCO~\cite{mscoco} is one of the most popular datasets containing 328k images with rich annotations.
Some other datasets include PASCAL VOC~\cite{pascal}, Visual Genome~\cite{genome}, Cityscapes~\cite{cityscapes}, MPII~\cite{mpii}, \etc.
Although these datasets facilitate the development of deep visual learning, they mainly serve for 2D single-view image understanding, which limits their applications in 3D vision. 

\paragraph{Video datasets.} 
Another line is video datasets.
Pioneering works construct the HMDB-51~\cite{hmdb51} and UCF-101~\cite{ucf101} dataset.
Afterward, the ActivityNet~\cite{activitynet} and Kinetics~\cite{kinetics} datasets are constructed of a larger scale and variation, of which the latter has collected 650k video clips that cover 700 classes.
Besides, some datasets are built for human pose estimation~\cite{jhmdb, 3dpw} and object detection/segmentation/tracking~\cite{davis, otb, bdd100k}.
IEMOCAP~\cite{iemocap} provides video data for the task of multimodal emotion recognition.
MSVD~\cite{msvd} and MSR-VTT~\cite{msr-vtt} annotate videos with extra captions.
Further, HowTo100M~\cite{howto100m} proposes a larger-scale one of 136 million samples from narrated instructional videos.
ActivityNet Captions~\cite{activitynet-captions} introduces the task of dense-captioning events and constructs a dataset with 20k videos. 
These datasets are primarily for video understanding which is different from our objective.

\paragraph{3D datasets.}~As the applications in 3D vision attract increasing attention, various 3D datasets are proposed.
~One line of works focuses on indoor scenes~\cite{nyu, sun3d, sun-rgbd, scenenn, s3dis, scannet, matterport3d}, where S3DIS~\cite{s3dis} and ScanNet~\cite{scannet} are two of most popular datasets. 
In addition, some works provide 3D object point clouds for contextual object surface reconstruction~\cite{scan2cad, 3d-front}.
3D outdoor scenes are scanned via LiDAR sensors~\cite{kitti, nuscenes, waymo, semantic3d} for autonomous driving.
ShapeNet~\cite{shapenet} and ModelNet~\cite{modelnet} are two object-centric datasets that provide rich 3D Computer-Aided Design (CAD) models for shape analysis, followed by similar datasets~\cite{xiang2016objectnet3d, koch2019abc}, which are usually low-quality, untextured, and have a domain gap with real-world objects.
Another line of works~\cite{singh2014bigbird, choi2016large, park2018photoshape, uy2019revisiting, collins2022abo,toscene} advocate real 3D objects but are still limited in scale.
We close this gap by shooting multi-view images of real-world objects, which capture the 3D awareness while allowing a scalable collection. 

\paragraph{Multi-view image datasets.} Multi-view image data is recently regarded as the source of 3D reconstruction or novel view synthesis.
Early works collect multi-view images of real objects but only provide 3D models that are approximated~\cite{pascal3d} or for only a few instances~\cite{choi2016large}. 
Henzler \etal~\cite{unsupervised_videos} contribute a larger video dataset to benchmark the task of 3D reconstruction. 
Another concurrent dataset, Objectron~\cite{objectron}, annotates 3D bounding boxes and ground planes for all objects but lacks camera poses or point clouds for all \~15k videos.
GSO~\cite{gso} gets clean 3D models with textures via scanning common household items but includes only a limited number of samples.
Some works also construct synthetic multi-view datasets~\cite{gao2022objectfolder, tremblay2022rtmv, shapenet, xie2020pix2vox++, hm3d-abo}.
Nevertheless, these datasets have a small scale and category range for special tasks, which limits the robust and generic learning of 3D deep models. 
We construct a large-scale multi-view image dataset -- MVImgNet, which contains 219k videos for real-world objects in a wide range of 238 categories.

\textbf{\textit{Special discussions on CO3D \cite{co3d}}.}
Very recently, CO3D extends \cite{unsupervised_videos} to 19k videos covering 50 object categories.
Although we share a similar idea in the light of object-centric multi-view data collections, our MVImgNet enjoys conspicuously larger scales.

More importantly, we have fundamentally different \textit{motivations} and \textit{insights}.
The primary motivation of CO3D is \textit{exactly pre-set}, which is to transfer the 3D reconstruction training/evaluation from synthetic datasets into the real-world setup.
In contrast, MVImgNet examines existing 3D datasets by re-walking the past development journey in the 2D domain.
Concretely, through reflecting on how ImageNet \cite{imagenet} demonstrates its generic impact on data-hungry algorithms,
we attempt to build a 3D counterpart of ImageNet and choose multi-view images as a soft bridge between 2D and 3D.
Instead of nailing down a special target, we conduct a variety of \textit{pilot} studies on how MVImgNet can benefit miscellaneous visual tasks, aiming to inspire and retain lots of possibilities for the broader vision community. 

In the later experiments, our datasets indeed show greater power than CO3D on different visual challenges.



\begin{figure}[t]
  \centering
   \includegraphics[width=0.45\textwidth]{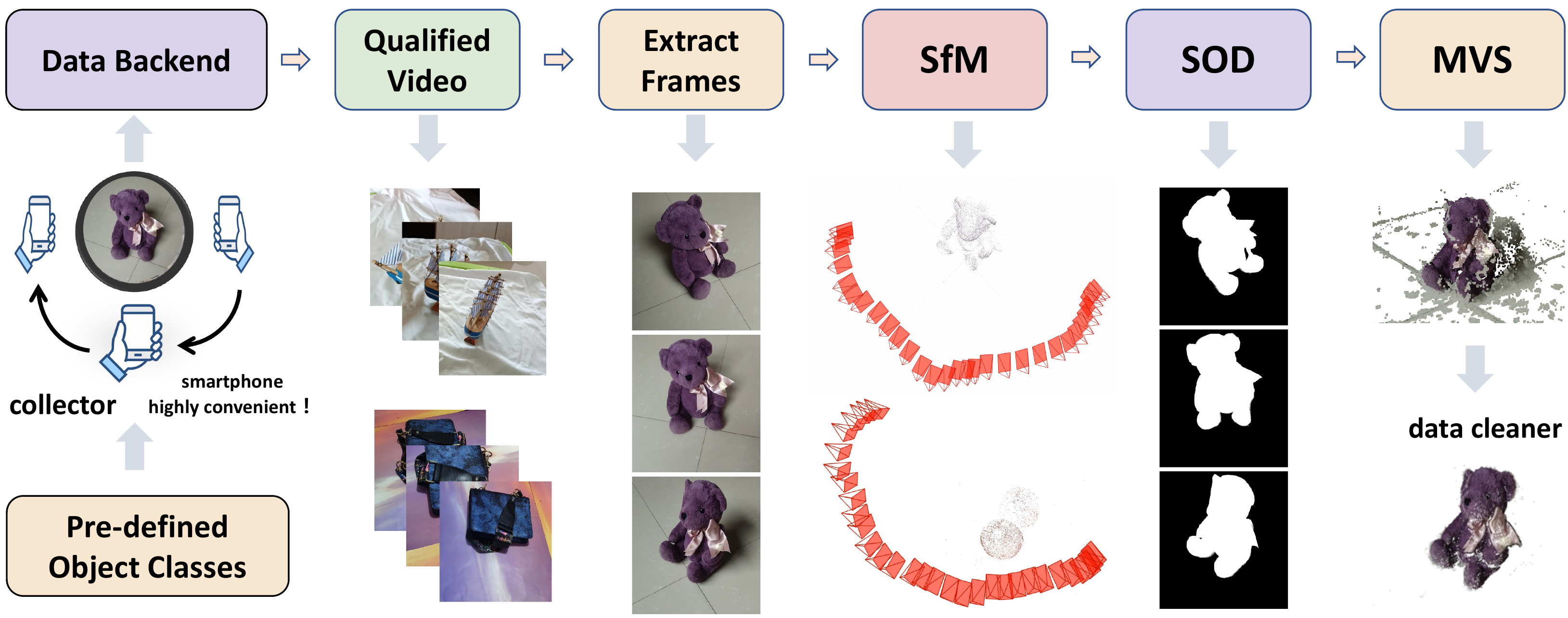}
   \caption{\textbf{The efficient data acquisition pipeline} of MVImgNet.}
   \label{fig:pipeline}
   \vspace{-0.4cm}
\end{figure}

\section{\textit{The Basis} -- MVImgNet Dataset}%
\label{sec:method}
As shown in Fig.~\ref{fig:pipeline}, the whole data acquisition pipeline of MVImgNet is highly efficient, which is illustrated below.

\subsection{Raw Data Preparation}
Building a large-scale dataset is always challenging, due to not only the laborious data collection but also the non-trivial annotation, which is especially critical for real-world 3D data \cite{scannet,s3dis,matterport3d}.
Thanks to the rapid development of mobile devices, shooting a video around an object in the wild becomes highly convenient and accessible in our daily life, which makes multi-view images be easily gained by crowd-sourcing.

%

\begin{table}[t]
  \centering
  \resizebox{0.55\linewidth}{!}{
  \begin{tabular}{cccc}
    \toprule
    Annotation  & Collected & Valid & GPU hours \\
    \midrule
    Sparse & 219,188 & 215,755 & 3,806.8 \\
    Segmentation & 104,261  & $\backslash$ & 2,316.9 \\
    Dense & 98,899  & 80,000 &   25,122.4\\
    \bottomrule
  \end{tabular}
  }
  \caption{\textbf{Data statistics}, including \textbf{collected} amount, \textbf{valid} amount after cleaning, and the \textbf{GPU hours} for processing.}
  \label{tab:statistic}
  \vspace{-0.2cm}
\end{table}

\begin{table}[t]
    \centering
    \resizebox{\linewidth}{!}{
    \begin{tabular}{l|c|c|c|c|c}
    \toprule
    Dataset & Real &\# of objects & \# of categories & Multi-view & 3D-GT\\
    \midrule
    ShapeNet ~\cite{shapenet} &\XSolid & 51k & 55& render& CAD model\\
    ModelNet~\cite{modelnet}  & \XSolid& 12k & 40 & render& CAD model\\
    Choi \etal~\cite{choi2016large} & \Checkmark & 2k & 9 & 360$^{\circ}$ captured & RGB-D scan\\
    Objectron ~\cite{objectron} & \Checkmark & 15k & 9 & limited & 3D bbox, pcl\\
    GoogleScan~\cite{gso} &\Checkmark & 2k & NA & 360$^{\circ}$ captured & RGB-D scan\\
    Henzler \etal~\cite{unsupervised_videos} &\Checkmark & 2k & 7 & 360$^{\circ}$ captured & pcl\\ScanObjectNN~\cite{scanobjectnn} & \Checkmark& 14k & 15 & limited & pcl\\
    CO3D~\cite{co3d} & \Checkmark& 19k & 50 & 360$^{\circ}$ captured& -\\
    CO3D-pcl~\cite{co3d} & \Checkmark& 5k & 50 & 360$^{\circ}$ captured & pcl\\
    \rowcolor{gray!20}
    MVImgNet (ours) & \Checkmark& 220k & 238 & 180$^{\circ}$ captured & -\\
    \rowcolor{gray!20}
    MVPNet (ours) & \Checkmark& 80k & 150 & 180$^{\circ}$ captured & pcl\\
    \bottomrule
    \end{tabular}
    }
    \caption{\textbf{Comparison} between our datasets and related ones. “pcl” denotes point clouds, “bbox” means bounding boxes.}
    \label{tab:multiview_comp}
    \vspace{-0.5cm}
\end{table}

\paragraph{Composition setup.}
We set up some constraints on the ratio.
Depending on the category, each class is initially set with the different expected number of video captures regarding their \textit{generality}, \eg, the number of captures for ``bottles'', ``bags'' and ``snacks'' is planned to be around 2000, while the number of ``chips'', ``apples'' and ``guitars'' is set to about 1000.
This setting is closer to real life.

\paragraph{Video capture.}
How to capture videos directly affect the quality of our data, so we draw up the following requirements as guidance for the captured videos:
1) The length of each video should be about 10 seconds. 
2) The frames in the video should not be blurred. 
3) Each video should capture $180^\circ$ or $360^\circ$ view of the object as much as possible. 
4) The proportion of the object in the video should be above 15\%. 
5) Each video can only contain one class of principal object. 
6) The captured object should be stereoscopic.

\paragraph{Crowdsourcing.}
We employ around 1000 normal collectors from different professions (\eg, teacher, doctor, student, cook, babysitter) and ages (20$\sim$50). 
Each of them is asked to take several videos in their daily life, (\ie, denoting diverse real-world environments) and upload them to the backend. 
Meanwhile, about 200 new well-trained expert data cleaners are responsible to review each submission and assure it fulfills the aforementioned capture requirements, when they may report some feedback or directly delete the unqualified submissions.
The whole procedure ensures both the \textit{diversity} and \textit{quality} of the raw videos.


%
%


\subsection{Data Processing}\label{sec:data_processing}
For each qualified video submission, we exploit an automatic process to obtain the common 2D and 3D annotations, including object masks, camera intrinsic and extrinsic, depth maps, and point clouds. 

\paragraph{Sparse reconstruction.}
Following the procedure of~\cite{llff, nerf}, the sparse reconstruction aims to reconstruct the camera intrinsic and extrinsic for each video, by applying the COLMAP Structure-from-Motion (SfM) algorithm~\cite{schoenberger2016sfm} on a series of equal-time-interval chosen frames. 

\paragraph{Foreground object segmentation.}
Each frame extracted from the original video is fed to the CarveKit~\cite{carvekit} for generating the binary foreground object mask, not only benefiting the dense reconstruction but also contributing to the further step of salient object detection (Sec.~\ref{sub:sod}). 

\begin{figure}[t]
  \centering
   \includegraphics[width=0.75\linewidth]{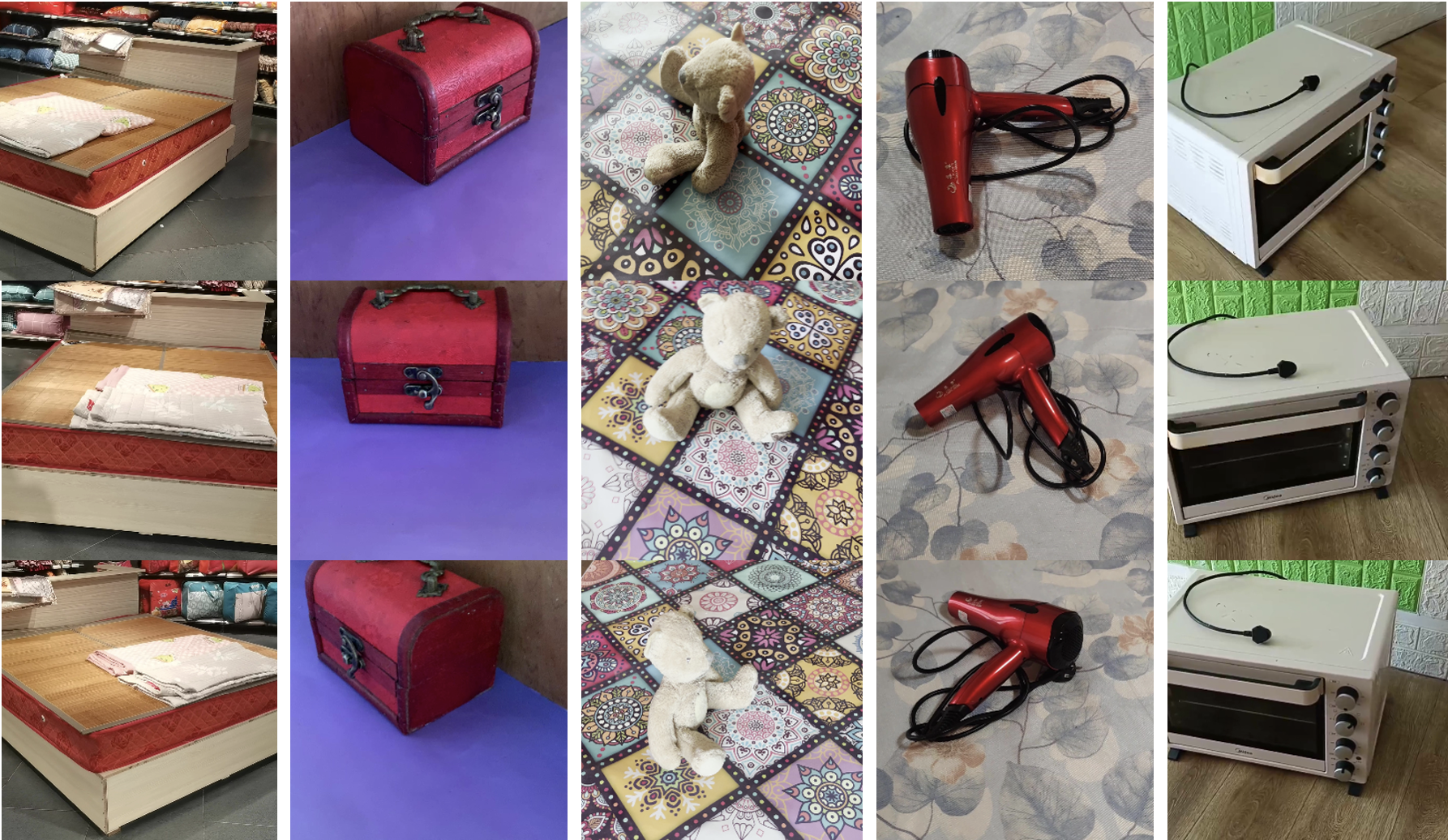}
   \caption{Some \textbf{frames} sampled from \textbf{MVImgNet}.} 
   \label{fig:pano}
   \vspace{-0.5cm}
\end{figure}

\paragraph{Dense reconstruction.}
With the sparse model output from COLMAP SfM, we employ multi-view stereo \cite{schoenberger2016mvs} of COLMAP to generate the dense depth and normal maps for each frame. We extract the depths of an object using the binary foreground masks, 
which are then back-projected and fused according to the normal information, yielding a densely reconstructed point cloud for each video.
Finally, the point cloud is manually cleaned by: 
1)~Delete the object with obvious noisy or extremely sparse reconstruction. 
2)~Remove all backgrounds.
The final derived 3D point cloud dataset -- MVPNet, is illustrated in Sec.~\ref{sec:MVPNet}.


\subsection{Dataset Summary}

\paragraph{Statistics.}
The statistics of MVImgNet are shown in \cref{tab:statistic}, and \cref{tab:multiview_comp} compares our datasets with other alternatives.
MVImgNet includes 238 object classes, from 6.5 million frames of 219,188 videos.
Fig.~\ref{fig:pano} shows some samples of MVImgNet.
The annotation comprehensively covers object masks, camera parameters, and point clouds.

\paragraph{Category.}
We leverage WordNet~\cite{wordnet} taxonomy that is used by ImageNet~\cite{imagenet} to describe multi-hierarchy categories of objects and define 238 common classes. 
Unlike ImageNet which contains various plants and animals (nature-centric), the objects in our MVImgNet are \textit{found} or \textit{used} in human daily life (human-centric), where 65 classes overlap with ImageNet. 
The detailed category taxonomy, per-category data distributions, and more sample visualizations of MVImgNet are illustrated in the supplementary material.




\begin{table*}[h]
 \begin{minipage}{0.62\linewidth}
 \centering
        \resizebox{0.99\linewidth}{!}{
	\begin{tabular}{l|cc
 c|ccc|ccc}
		\toprule
		& \multicolumn{3}{c|}{Diffuse Synthetic $360^\circ$~\cite{sitzmann2019deepvoxels}} & \multicolumn{3}{c|}{Realistic Synthetic $360^\circ$~\cite{nerf}} & \multicolumn{3}{c}{Real-world $360^\circ$ Objects~\cite{co3d}} \\
		
Method & PSNR$\uparrow$ & SSIM$\uparrow$ & LPIPS$\downarrow$ & PSNR$\uparrow$ & SSIM$\uparrow$ & LPIPS$\downarrow$ & PSNR$\uparrow$ & SSIM$\uparrow$ & LPIPS$\downarrow$ \\
		\midrule
		Train from scratch & 37.17 & 0.990 & 0.017 & 25.49 & 0.916 & 0.100 & 22.18 & 0.714 & 0.365\\
  MVImgNet-pretrained & \textbf{37.66} & \textbf{0.990} & \textbf{0.014} & \textbf{27.26} & \textbf{0.930} & \textbf{0.071} & \textbf{24.67} & \textbf{0.740} & \textbf{0.310}\\
		\bottomrule
	\end{tabular}
 }
	\caption{\textbf{NeRF} quantitative results on three different levels of test sets.}
  \label{tab:nerf}
  \vspace{-0.2cm}
 \end{minipage}
 \begin{minipage}{0.41\linewidth}
 \centering
        \resizebox{0.66\linewidth}{!}{
	\begin{tabular}{l|ccc}
		\toprule
		& \multicolumn{3}{c}{Real-world $360^\circ$ Objects \cite{co3d}} \\
		Where to pretrain & PSNR$\uparrow$ & SSIM$\uparrow$ & LPIPS$\downarrow$ \\
		\midrule
		CO3D \cite{co3d} & 24.01 & 0.732 & 0.339\\
		MVImgNet-\textit{small} & 24.08 & 0.736 & 0.316\\
		MVImgNet & \textbf{24.67} & \textbf{0.740} & \textbf{0.310}\\
		\bottomrule
	\end{tabular}
 }
\caption{\textbf{NeRF} quantitative comparison with CO3D \cite{co3d}.}
 \label{tab:nerf_co3d}
  \vspace{-0.2cm}
  \end{minipage}  
 \vspace{-0.4cm}
\end{table*}

\section{3D Reconstruction}%
\label{sec:3d_recon}

\subsection{Radiance Field Reconstruction}%
\label{sub:nerf}

\newcommand{\gtfigwidth}{0.15\linewidth}
\newcommand{\resultswidth}{0.18\linewidth}

\newcommand{\insertimg}[1]{
  \makecell{
  \includegraphics[width=\resultswidth]{#1} \\
  }
}

\begin{figure}[t]
	\centering
	\scriptsize
	\begin{tabular}{@{}c@{\,\,}c@{}c@{}c@{}c@{}}
		\makecell[c]{
			\includegraphics[trim={0px 0px 0px 0px}, clip, width=\gtfigwidth]{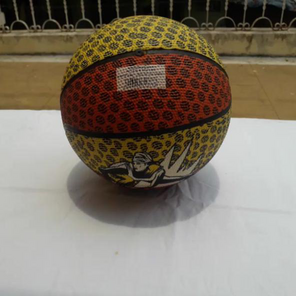}
			\\
			\textit{Ball}
		}
		& 
		\insertimg{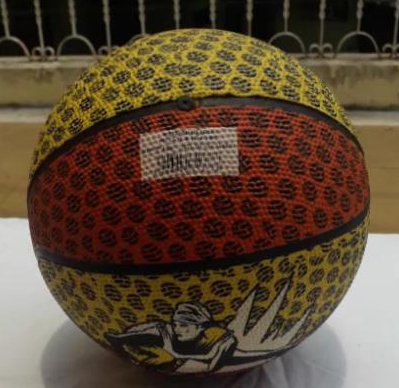} &
		\insertimg{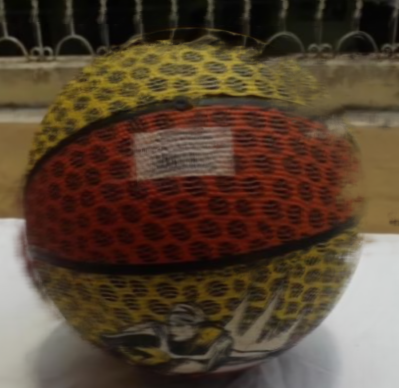} &
		\insertimg{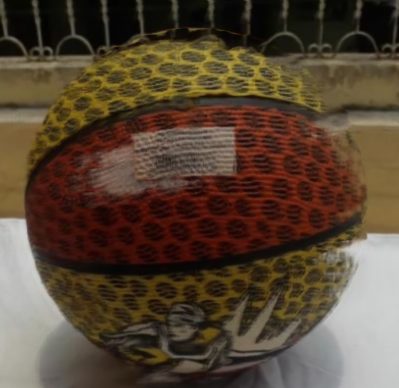} &
		\insertimg{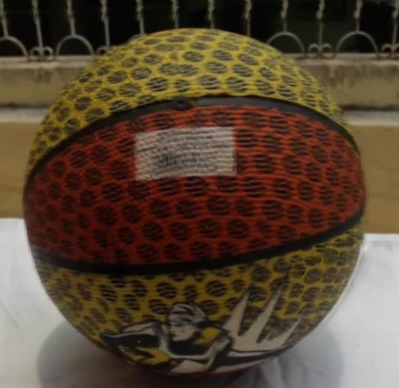} \\
		\makecell[c]{
			\includegraphics[trim={0px 0px 0px 0px}, clip, width=\gtfigwidth]{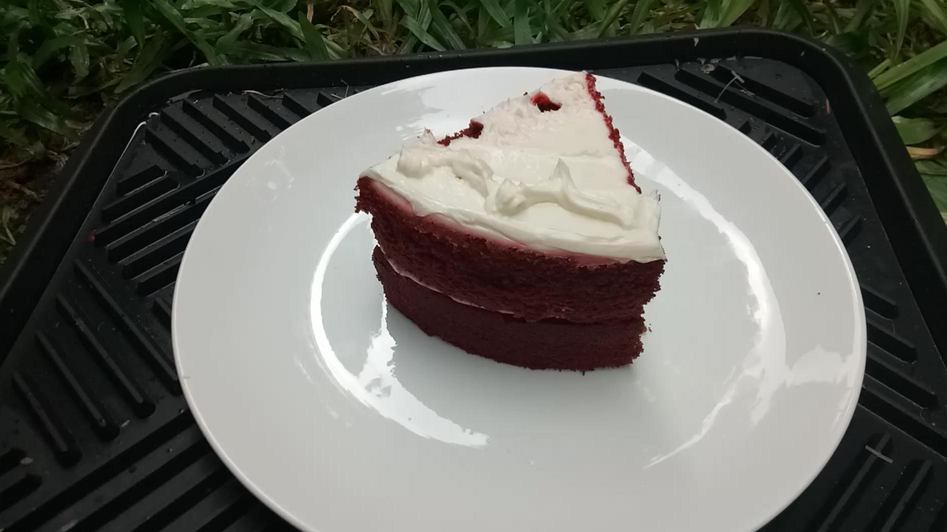}
			\\
			\textit{Book}
		}
		& 
		\insertimg{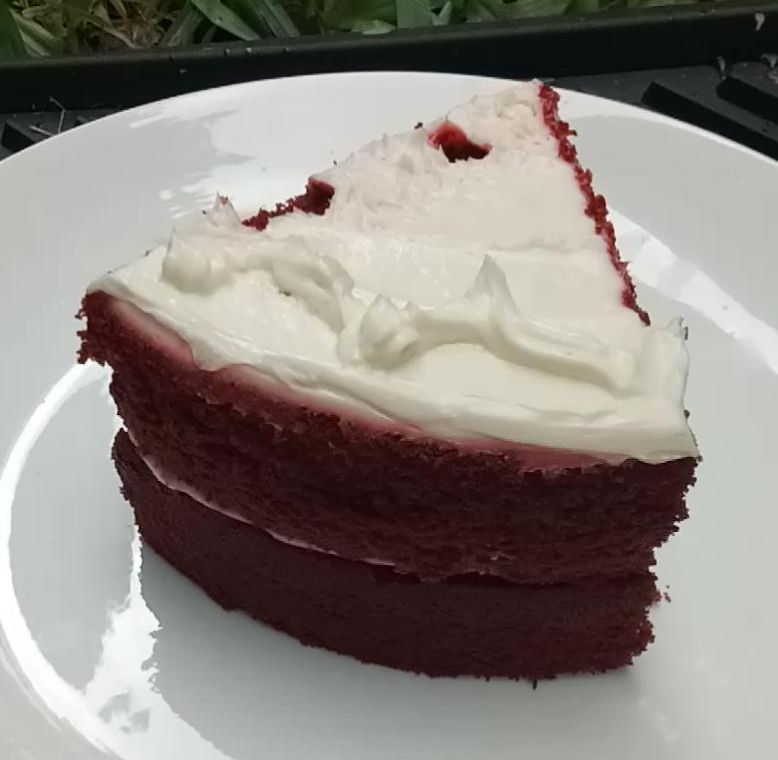} &
		\insertimg{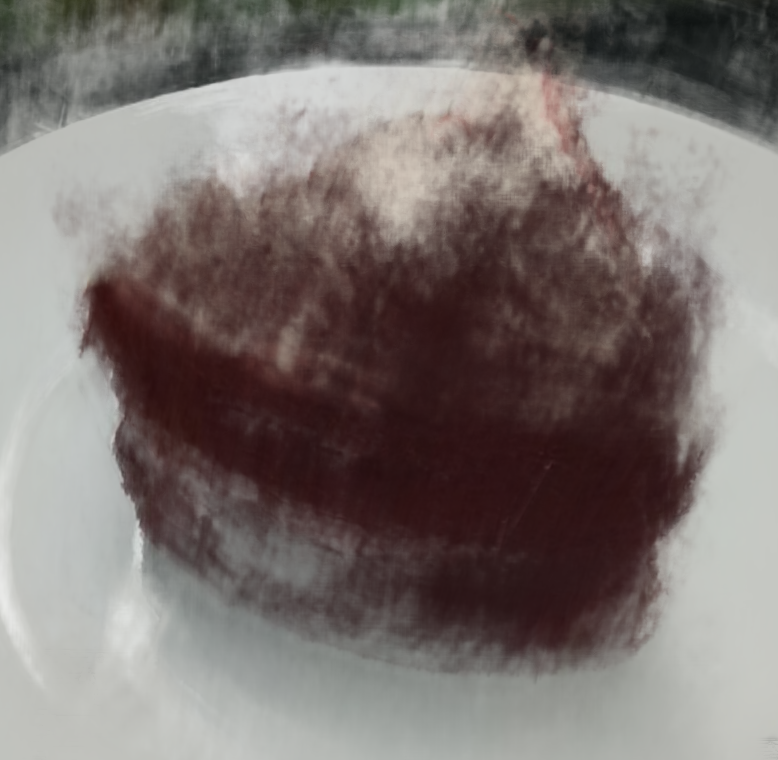} &
		\insertimg{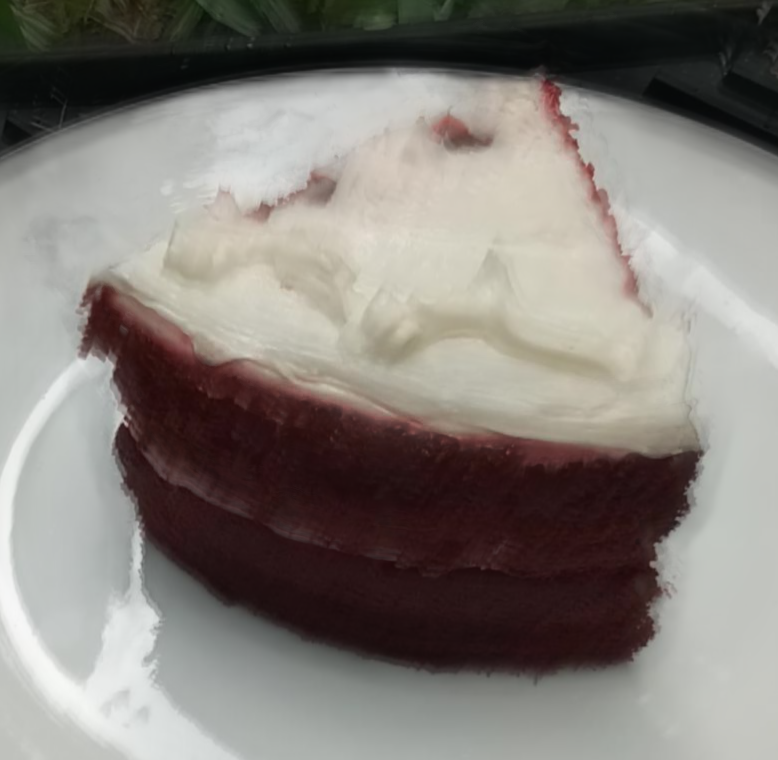} &
		\insertimg{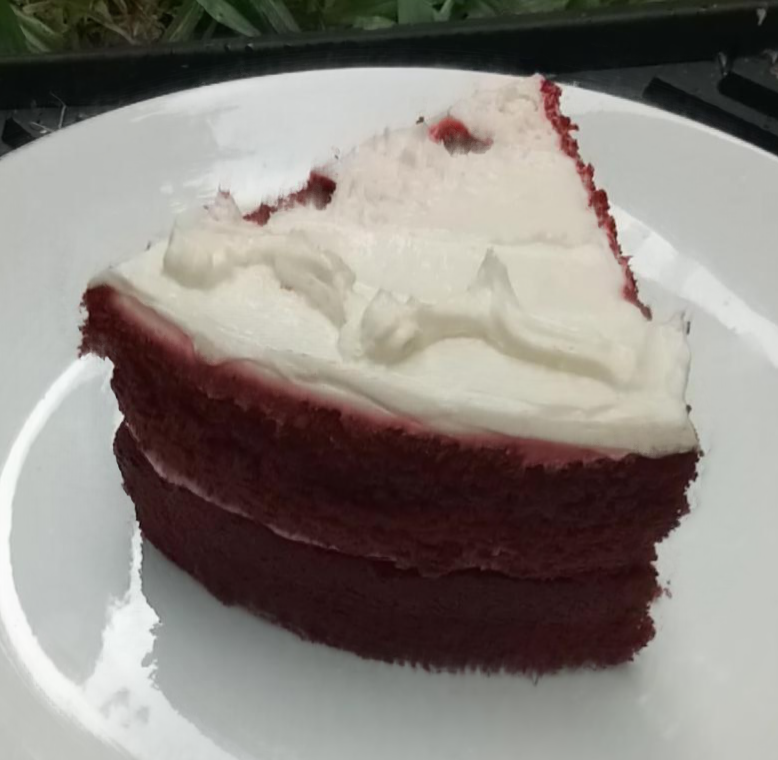} \\
		& GT  &  Scratch & CO3D-PreTr. & MVImgNet-PreTr.
	\end{tabular}
	\caption{\textbf{NeRF} qualitative comparison of \textbf{train-from-scratch} IBRNet, CO3D~\cite{co3d}-pretrained IBRNet and \textbf{MVImgNet-pretrained} IBRNet~\cite{wang2021ibrnet}.
	}
	\label{fig:nerf}
 \vspace{-0.4cm}
\end{figure}

\paragraph{Pre-review.}
Recently, a series of generalizable Neural Radiance Fields (NeRF) variants~\cite{yu2021pixelnerf, wang2021ibrnet, trevithick2021grf, chen2021mvsnerf} have been proposed to reconstruct radiance field on-the-fly from one/few-shot source views for novel view synthesis.

\paragraph{What can MVImgNet do?}
Training NeRFs that can generalize to unseen objects requires learning 3D priors from a huge amount of multi-view images.
Existing state-of-the-arts either resort to learning on large-scale synthetic data~\cite{yu2021pixelnerf, trevithick2021grf, chen2021mvsnerf}, or adopt a mixed use of synthetic data and a small self-collected real dataset~\cite{wang2021ibrnet}. 
However, the synthetic data introduces a big domain gap with real-world objects, only a few real scenes cannot remedy this defect.
We argue that: \textit{our MVImgNet perfectly fits the huge data demand of learning-based generalizable NeRF methods.}

To verify this, we choose IBRNet~\cite{wang2021ibrnet} as the baseline and conduct an empirical study.
We pretrain IBRNet on the full MVImgNet dataset then finetune on the training datasets used in IBRNet \cite{wang2021ibrnet} for a few iterations, and compare with the original train-from-scratch IBRNet model in terms of generalization capability (\ie, generalizing to unseen scenes with only few-shot inputs). 
For fairness, we evaluate all methods under the same protocol of IBRNet.

Here comes a challenge about \textit{how to evaluate} the generalization ability of NeRFs, since there is no official benchmark for this.
To this end, we employ three different \textit{third-party} object-centric datasets to form the test set. 1) The diffuse synthetic 360$^\circ$ object dataset~\cite{sitzmann2019deepvoxels} which contains $4$ Lambertian objects. 2) The realistic synthetic 360$^\circ$ object dataset~\cite{nerf} which consists of $8$ realistically fabricated objects. 3) The real-world 360$^\circ$ object dataset~\cite{co3d} which includes $88$ real-world scenes from different lighting conditions.
To summarize, the whole test set comprises $100$ objects from $56$ distinct categories, ranging from the synthetic domain to the real-world domain, which is considered to be an impartial evaluation set for generalization ability.

The quantitative and qualitative results are respectively shown in \cref{tab:nerf} and \cref{fig:nerf}. Evidently, pretraining on MVImgNet improves the generalization ability of the model by a large margin.
Moreover, we perform the same pretraining on CO3D \cite{co3d} and MVImgNet-\textit{small} (a random subset of MVImgNet, which owns the same scale as CO3D). 
As shown in \cref{tab:nerf_co3d} and \cref{fig:nerf}, MVImgNet shows greater power than CO3D \cite{co3d}.
The implementation details can be found in the supplementary material.

\begin{figure}[t]
  \centering
   \includegraphics[width=0.7\linewidth]{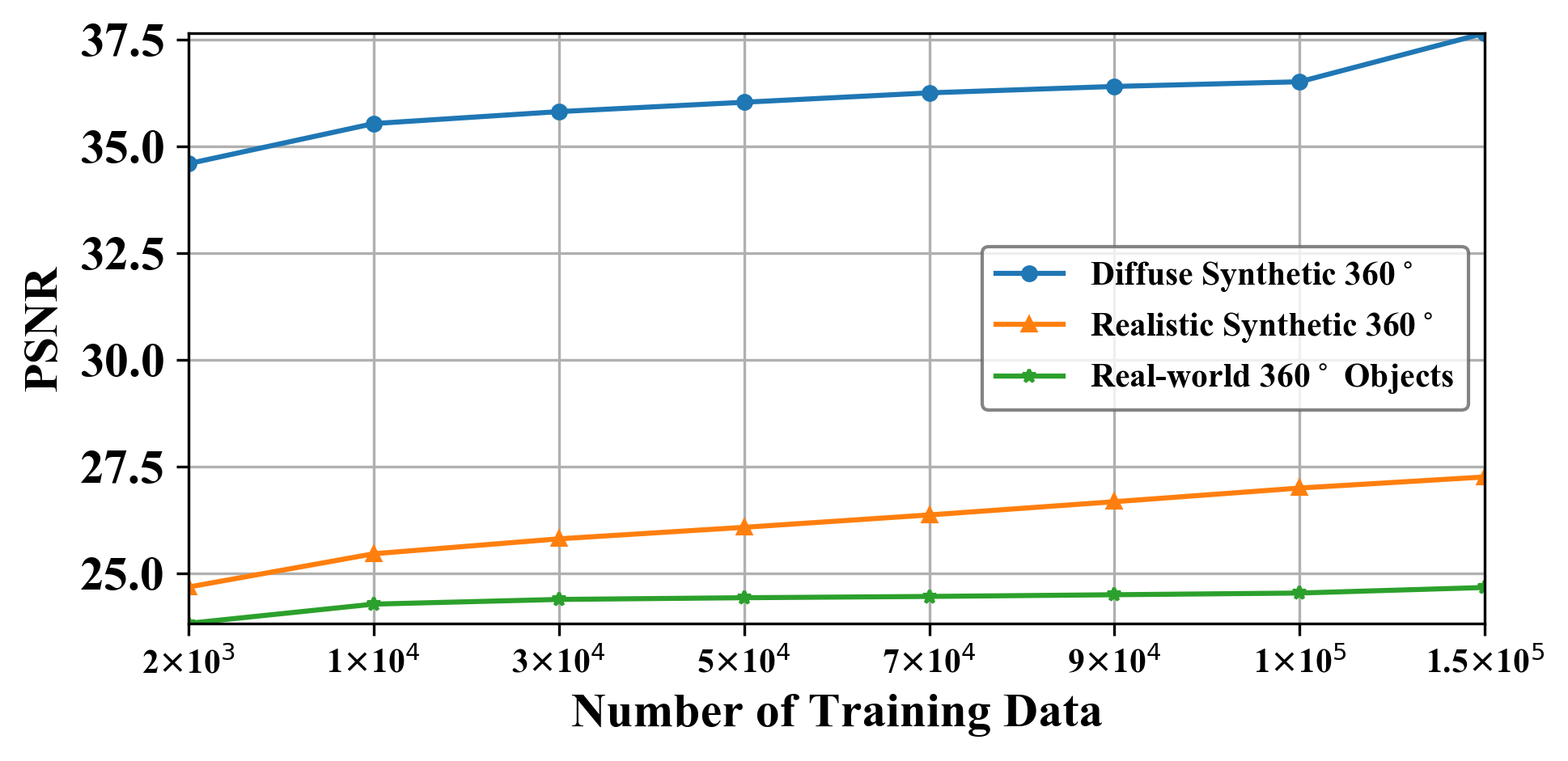}
   \caption{PSNR of IBRNet pretrained by \textbf{a different number of MVImgNet data} (higher is better).}
   \label{fig:nerf_scala}
   \vspace{-0.4cm}
\end{figure}

\paragraph{More data, more power.}
\cref{fig:nerf_scala} shows that an apparent rising trend of generalization metrics can be observed with the increase of training data. 


\subsection{Multi-view Stereo}%
\label{sub:mvs}

\paragraph{Pre-review.}
Multi-view Stereo (MVS) \cite{MVS} is a classical task in 3D computer vision, with the goal of reconstructing 3D geometry from multi-view images.
Conventional methods \cite{vbrs,hu2012least,amv,pmb} reconstruct 3D geometry by finding the patches matched with different images and estimating the depth according to the extrinsic camera.
In recent years, deep learning methods are introduced into MVS to solve the issues such as weak textures, and non-laplacian spheres.
Two popular end-to-end methods MVSNet \cite{yao2018mvsnet} and R-MVSNet \cite{yao2019recurrent} propose to encode the multi-view images and build a cost volume for predicting depth maps. 
Under the supervision of large-scale RGB-D datasets \cite{dtu}, they outperform conventional patch-matched-based approaches.
However, these methods require massive amounts of RGB-D data, which is always difficult to acquire.
This drives the emergence of self-supervised MVS methods \cite{xu2021self, chang2022rc, ding2022kd, yang2021self}.

\paragraph{What can MVImgNet do?}
We demonstrate that our MVImgNet is capable of benefiting the \textit{data-efficient} MVS with limited training examples, which is practically meaningful considering MVS always requires the burdensome collection of RGB-D data.

We pretrain a self-supervised MVS method, JDACS \cite{xu2021self}, on MVImgNet.
Our implementation follows the original settings of JDACS.
Then, we select DTU \cite{dtu} dataset with 79 training samples and 22 test samples. 
We perform the \textit{data-efficient} evaluation with limited training data, where the MVImgNet-pretrained MVSNet is finetuned on 5\%, 15\%, and 25\% of DTU training samples, and evaluated on the DTU test set.
\cref{tab:mvs} reports the accuracy given different thresholds of the error between the predicted and ground truth depth map, which indicates that the MVImgNet-pretrained model is capable of improving the model trained from scratch by a large margin under the data-efficient setup.
Furthermore, we perform pretraining using CO3D \cite{co3d} and MVImgNet-\textit{small}. \cref{tab:mvs_co3d} indicates that our MVImgNet is stronger than CO3D, and MVImgNet-\textit{small} shows comparable power as CO3D.
For more implementation details, please see the supplementary material.


\begin{table}[t]
	\centering
        \resizebox{0.65\linewidth}{!}{
	\begin{tabular}{l|ccc}
\toprule
Ratio & $2mm\uparrow$ & $4mm\uparrow$ & $8mm\uparrow$  \\ 
\midrule
$5\%$  & 48.61 / \textbf{50.10}  & 65.18 / \textbf{67.97} & 75.69 / \textbf{78.58}   \\ 
$15\%$ &  \textbf{54.74} / 54.68 & 70.96 / \textbf{72.04} & 80.32 / \textbf{81.73} \\ 
$25\%$ &   57.29 / \textbf{58.63 } &  74.41  / \textbf{75.20} &  83.68 / \textbf{84.28} \\   \bottomrule
\end{tabular}
}
	\caption{\textbf{MVS depth map accuracy} on DTU evaluation set under \textbf{different ratios} of DTU training samples, in terms of \textbf{training from scratch / MVImgNet-pretrained} (higher is better).}
	\label{tab:mvs} 
 \vspace{-0.2cm}
\end{table}

\begin{table}[t]
	\centering
        \resizebox{0.62\linewidth}{!}{
	\begin{tabular}{l|ccc}
\toprule
Where to pretrain & $2mm\uparrow$ & $4mm\uparrow$ &  $8mm\uparrow$  \\ 
\midrule
CO3D & 46.18 & 66.72 & 78.39  \\
MVImgNet-\textit{small} & 47.73 & 66.53 & 78.15 \\ 
MVImgNet  & \textbf{50.10}  & \textbf{67.97} &  \textbf{78.58} \\
\bottomrule
\end{tabular}
}
	\caption{\textbf{MVS depth map accuracy} comparison with CO3D \cite{co3d} on DTU evaluation set. Pretrain on MVImgNet, MVImgNet-\textit{small} or CO3D, and finetune using $5\%$ of DTU training samples.} 
	\label{tab:mvs_co3d} 
\vspace{-0.4cm}
\end{table}

  	

\textbf{\textit{We advocate benchmarking NeRF and MVS methods with the help of pretraining on MVImgNet.}}

\section{View-consistent Image Understanding}%
\label{sec:recongnition}


\subsection{View-consistent Image Classification}
\label{sub:img_cls}

\noindent\textbf{Pre-review.}
As indicated by Dong \etal \cite{dong2022viewfool}, although humans can easily recognize one object from different views, deep learning models can hardly do that robustly. 
MVImgNet provides numerous images from different viewpoints, so we hope to enhance the model's view consistency with the help of MVImgNet, which is significantly important for realizing human-like intelligence.

\noindent\textbf{What can MVImgNet do?}
One naive approach is to finetune the ImageNet-pretrained ResNet-50 \cite{he2016deep} on our MVImgNet. 
However, such an approach will be problematic due to the categories of two datasets are very distinct, which may cause catastrophic forgetting issues \cite{kirkpatrick2017overcoming}. 

For this reason, we create a \textit{new training set}, namely \textbf{MVI-Mix}, by mixing the original ImageNet data with MVImgNet. 
Specifically, we randomly sample 5 \textit{consecutive} frames of each video in MVImgNet, and mix the multi-view images with original ImageNet data.
We also build another two artificial training sets, namely MVI-Gap and MVI-Aug, that only differ MVI-Mix with image views.
\textbf{MVI-Gap} samples the 5 \textit{non-consecutive} frames that have much larger view differences.
\textbf{MVI-Aug} samples 1 frame from each video of MVI-Mix and applies \textit{4 different data augmentations} (\ie, random color jittering, grid mask, rotation and erase, respectively) to it, which aims to differentiate the multi-view augmentation from the normal data augmentations.
We compare the \textit{variance} of the softmax confidence and the accuracy on MVImgNet \textit{test set} for examining the view consistency.
As illustrated in \cref{tab:cls_res}, adding our MVImgNet data for training can effectively improve the model's view consistency and accuracy. The test dataset used for all experiments is MVImgNet test dataset. In addition to using the convolution-based network (\ie, ResNet-50), we also apply the \textit{Transformer}-based architecture, DeiT-Tiny \cite{deit}.
The results are 48.76\% accuracy and 0.225 Var on ImageNet-only, VS 0.122 Var (\textbf{0.103}$\downarrow$) and 73.88\% (\textbf{25.12}$\uparrow$) accuracy on MVI-Mix. This leads to a unanimous conclusion.

We further construct CO3DI-Mix (corresponding to MVI-Mix) as the mixer of CO3D \cite{co3d} \& ImageNet under the same ratio of MVImgNet \& ImageNet in MVI-Mix. As illustrated in \cref{tab:co3d-mix-comp}, MVImgNet brings more benefits than CO3D for view-consistent image recognition.

\begin{table}
  \centering
  \resizebox{0.56\linewidth}{!}{
  \begin{tabular}{lcc}
    \toprule
    Dataset  & Confidence Var  & Accuracy \\
    \midrule
    ImageNet-only      &0.207      & 53.09\% \\
    MVI-Aug     & 0.105  & 71.48\% \\
    MVI-Gap     & 0.103  & 77.23\% \\
    MVI-Mix    & \textbf{0.102}  & \textbf{77.31}\% \\
    \bottomrule
  \end{tabular}
  }
  \caption{\textbf{View-consistent image classification results} on MVImgNet test set using \textbf{fully-supervised ResNet-50 \cite{he2016deep}}. \textbf{Adding MVImgNet for training improves the view consistency} (smaller Var and higher Acc indicate better view consistency).}
  \label{tab:cls_res}
  \vspace{-0.2cm}
\end{table}

\begin{table}[t] 
\centering
\resizebox{0.7\linewidth}{!}{
    \begin{tabular}{@{}l|ll@{}}
    \toprule
                  & Train: CO3DI-Mix                                                   & Train: MVI-Mix                                                     \\ \midrule
    Test: CO3D    & \begin{tabular}[c]{@{}l@{}}Var: 0.104 \\ Acc: 91.25\%  \end{tabular} & \begin{tabular}[c]{@{}l@{}}Var: 0.155 \textcolor{red}{(+0.051)}\\ Acc: 84.83\% \textcolor{red}{(-6.42\% )}\end{tabular} \\

    \midrule
    
    Test:MVImgNet & \begin{tabular}[c]{@{}l@{}}Var: 0.188 \textcolor{blue}{(+0.086)}\\ Acc: 57.93\% \textcolor{blue}{(-19.38\% )} \end{tabular} & \begin{tabular}[c]{@{}l@{}}Var: 0.102\\ Test: 77.31\%\end{tabular} \\ \bottomrule
    \end{tabular}
}

\caption{\textbf{View-consistent image classification} comparison with CO3D \cite{co3d}. 
The \textit{performance drops} when train on CO3DI-Mix and test on MVImgNet (\textcolor{blue}{blue number}) are much larger than the opposite (\textcolor{red}{red number}), which means that MVI-Mix pretrained model is more robust to multiview consistency than CO3DI-Mix pretrained model.}
\label{tab:co3d-mix-comp}
\vspace{-0.4cm}
\end{table}

\subsection{View-consistent Contrastive Learning}%
\label{sub:contras}

\noindent \textbf{Pre-review.}
Contrastive Learning~(CL) is one mainstream of self-supervised training techniques \cite{moco,byol,chen2020mocov2,chen2020contrastive,chen2020simsiam,chen2021mocov3,sim}.
One key factor of CL is the construction of positive/negative pairs, \eg, MoCo-v2~\cite{chen2020mocov2} treats an image with different augmentations, \textit{a.k.a.} \textbf{views}, as the positive pair.


\noindent \textbf{What can MVImgNet do?}
One natural question is: \textit{can the viewpoints of MVImgNet serve as the positive pairs for CL?}
To answer this question, we finetune the off-the-shelf ImageNet-pretrained MoCo-v2 on MVImgNet.
For each iteration, we randomly sample two frames from the same video as the positive pair and apply the original data augmentations used in MoCo-v2 \cite{chen2020mocov2}.
Meanwhile, the frames from other videos will be treated as their negatives.
The MVImgNet-finetuned model is finally evaluated on the MVImgNet test dataset for examining the view consistency, where we compare the variance and mean of the softmax confidence and the accuracy. 
As \cref{tab:cls_pretrain} shows, finetuning on MVImgNet with CL can also improve both the model's view consistency and accuracy. 
More implementation details of view-consistent image classification is described in the supplementary material.

\textbf{\textit{In the future, it is highly recommended to regard view-consistency as a criterion for evaluating the image recognition task, and utilize MVImgNet to pretrain the models.}}


\begin{table}[t]
  \centering
  \scriptsize
  \begin{tabular}{lccc}
    \toprule
    Training scheme  & Confidence Var & Accuracy \\
    \midrule
    w/o finetune     & 0.098   & 70.26\% \\
    w/ finetune    & \textbf{0.086}   & \textbf{71.22}\% \\
    \bottomrule
  \end{tabular}
  \caption{\textbf{View-consistency classification results} under the \textbf{self-supervised contrastive learning} regime on MVImgNet test set. \textbf{Finetuning MoCo-v2 \cite{chen2020mocov2} on MVImgNet improves the view consistency}.}
  \label{tab:cls_pretrain}
  \vspace{-0.2cm}
\end{table}


\subsection{View-consistent SOD}%

\begin{figure}
  \centering

   \includegraphics[width=0.9\linewidth]{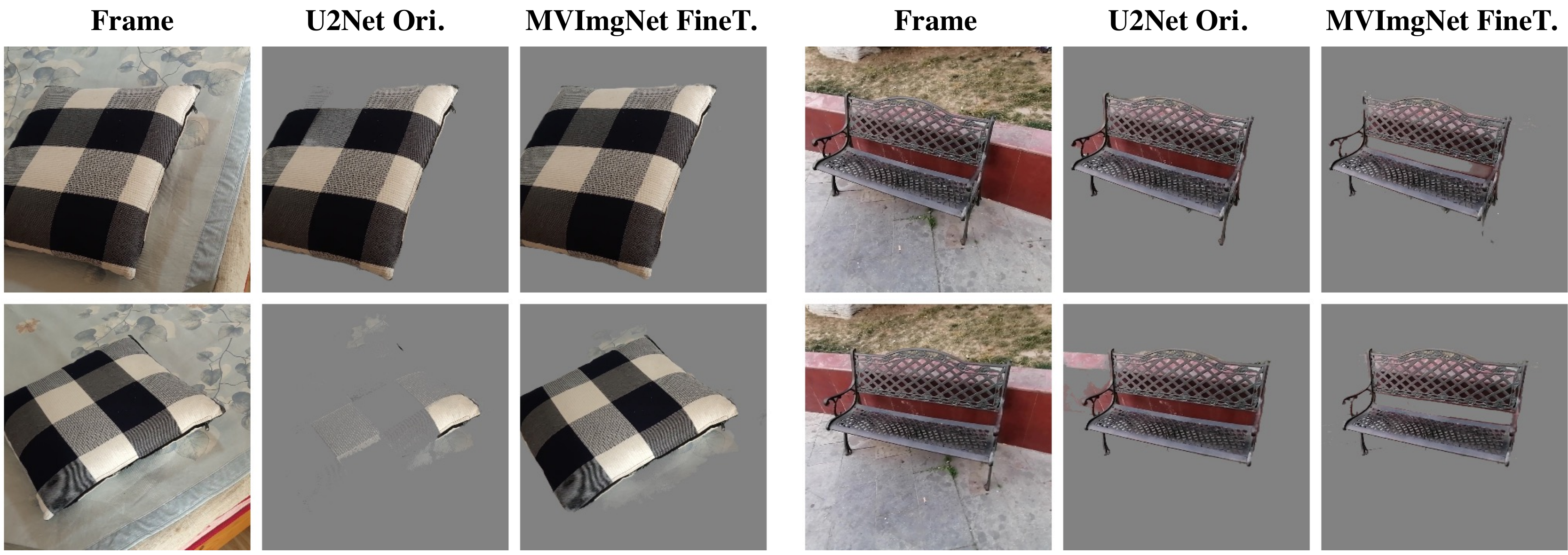}
   \caption{
    Qualitative comparison between \textbf{MVImgNet-finetuned} U2Net \cite{Qin_2020_PR} and \textbf{original} U2Net for salient object detection (SOD). Left: \textbf{finetuning on MVImgNet} improves the performance on a \textit{hard} view. Right: \textbf{finetuning on MVImgNet} improves the performance on two consecutive hard views.
   }
   \vspace{-0.4cm}
   \label{fig:sod_res}
\end{figure}
\noindent \textbf{Pre-review.}
Salient Object Detection (SOD) aims to segment the most visually prominent objects in an image. Although remarkable progress has been made recently, it remains lots of challenges.



\noindent \textbf{What can MVImgNet do?}
We test a state-of-the-art SOD model U2Net \cite{Qin_2020_PR} on our MVImgNet. 
As Fig.~\ref{fig:sod_res} shows, U2Net failed to segment ``hard" views, even though some views can be segmented with few flaws. 
Despite such a disappointment, the inconsistent predictions of different views caught our attention: \textit{can we improve the SOD models with multi-view consistency?}

We propose to leverage the multi-view consistency to improve SOD with the help of optical flows.
Specifically, given two consecutive frames, we first calculate their optical flow and warp the flow to the segmentation mask of one of the frames. 
Then, a consistency loss can be calculated between the warped mask and the mask of another frame. 
For ease of implementation, we directly finetune U2Net on our MVImgNet.
To prevent the catastrophic forgetting issue \cite{kirkpatrick2017overcoming}, in addition to the MVImgNet, we also use the original training data DUTS-TR\cite{dso}. For the ``hard" views (IoU $\leq 0.7$) on our MVImgNet test set, the model finetuned on our MVImgNet can bring a 4.1\% IoU improvement (see Fig.~\ref{fig:sod_res} for qualitative comparison).
The implementation details of view-consistent SOD can be found in the supplementary material.

\label{sub:sod}

\section{MVPNet for 3D Understanding}%
\label{sec:3d_understand}


\begin{figure}[t]
  \centering
  \includegraphics[width=0.8\linewidth]{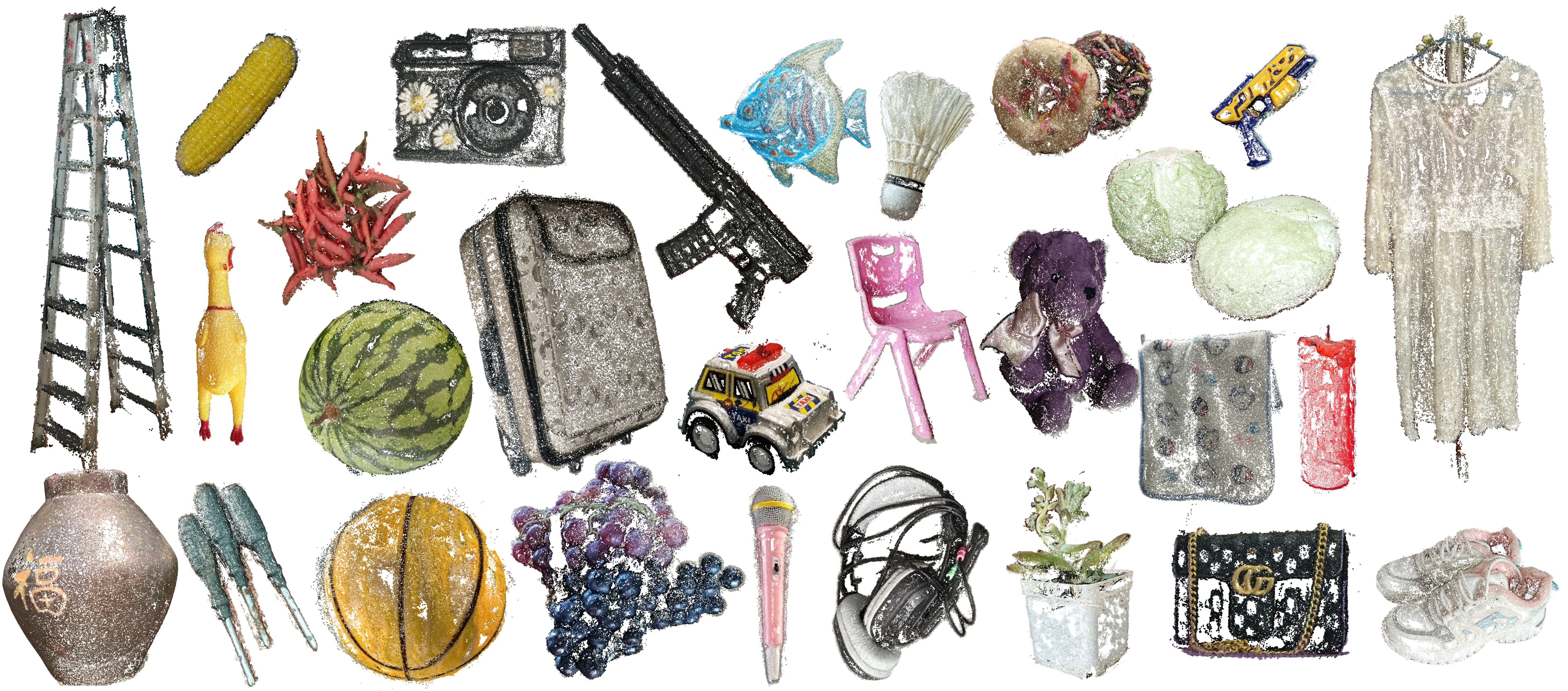}
   \caption{\label{fig:mvppano}Some \textbf{3D point clouds} sampled from \textbf{MVPNet}.}
   \vspace{-0.4cm}
\end{figure}

\subsection{MVPNet Dataset}%
\label{sec:MVPNet}


Derived from the dense reconstruction on MVImgNet (as mentioned in Sec. \ref{sec:data_processing}), a new large-scale real-world 3D object point cloud dataset -- MVPNet, is born, which contains 87,200 point clouds with 150 categories. 
\cref{fig:mvppano} shows some examples of MVPNet.
As listed in \cref{tab:multiview_comp}, compared with existing 3D object datasets, our MVPNet contains a conspicuously richer amount of real-world object point clouds, with abundant categories covering many common objects in the real life. 
The detailed category taxonomy per-category data distributions of MVPNet are illustrated in the supplementary material.

\subsection{3D Point Cloud Classification}
\label{sub:pointclass}

In this work, we focus on point cloud classification.
We believe that future works may also utilize our dataset to help much more 3D understanding tasks such as indoor-scene parsing, outdoor-environment perception, pose estimation, and robotics manipulation.

\paragraph{Pre-review.}
We advocate paying more attention to \textit{real-world} setup, which is more feasible for real applications.
ScanObjectNN \cite{uy2019revisiting} has been manifested as the most challenging point cloud classification benchmark so far, so we choose it for the major comparison with our MVPNet.

\paragraph{What can MVPNet do?} 
We show that pretraining on MVPNet is able to aid the performance of real-world point cloud classification.
We pretrain several models \cite{qi2017pointnet,qi2017pointnet++,wang2019dynamic,ma2022rethinking,xiang2021walk,xu2021learning,xu2021paconv,guo2021pct} on MVPNet, and finetune them on ScanObjectNN. 

Two settings are considered for evaluation.
First is PB\_T50\_RS in ScanObjectNN with small perturbation, translation and rotation on point cloud. 
Another is adding heavy rotation on PB\_T50\_RS, to create a more challenging setting. 
The results are shown in \cref{tab:quan_comp_pcd}, where pretraining on MVPNet is able to increase the classification accuracy under most circumstances.

\begin{table} \centering
\resizebox{0.95\linewidth}{!}{
    \begin{tabular}{l|c|c||c|c}
        \toprule
        &\multicolumn{2}{c||}{Add Random Rotation}&\multicolumn{2}{c}{PB\_T50\_RS}\\
       Method & from scratch & pretrained & from scratch & pretrained\\
        \midrule
        PointNet~\cite{qi2017pointnet} & 60.57 / 55.20 & \textbf{64.25 / 59.29} & \textbf{70.63 / 67.28} & 67.73 / 64.12\\
        PointNet++~\cite{qi2017pointnet++} & 76.50 / 73.42 & \textbf{78.76 / 76.54} & 78.80 / 75.70& \textbf{80.22 / 76.91}\\
        DGCNN~\cite{wang2019dynamic} & 80.50 / 78.45 & 80.42 / 78.20 & 79.44 / 76.24 & \textbf{82.36 / 80.08}\\
        PointMLP~\cite{ma2022rethinking} & 83.69 / 82.54 & \textbf{84.87 / 83.71} & 85.64 / 84.14 & \textbf{85.98 / 84.38}\\
        CurveNet~\cite{xiang2021walk} & 73.96 / 69.96& \textbf{78.99 / 76.59} & 74.27 / 69.43 & \textbf{83.68 / 81.17}\\
        GDANet~\cite{xu2021learning} & 80.33 / 79.14 &\textbf{83.59 / 82.29}&79.01 / 75.91&\textbf{83.90 / 82.51}\\
        PAConv~\cite{xu2021paconv} & 70.91 / 65.70&\textbf{76.21 / 72.47}&72.88 / 68.60z&\textbf{76.91 / 73.45}\\
        PCT~\cite{guo2021pct} & 81.32 / 79.17& \textbf{82.08 / 80.41} & 77.46 / 73.64& \textbf{84.20 / 81.94}\\
        \midrule
        PointMAE ~\cite{pointmae}  &  83.17 / 80.75 & \textbf{86.19 / 84.60} & 77.34 / 73.52 & \textbf{84.13 / 81.92}\\
        \bottomrule
    \end{tabular}
}
\caption{\label{tab:quan_comp_pcd} \textbf{ScanObjectNN real-world point cloud classification results}. The comparison is between the \textbf{train-from-scratch} model and \textbf{MVPNet pretrained} model. The metric is \textbf{overall / average accuracy}.}
\vspace{-0.1cm}
\end{table}

\begin{table}[t]
\centering
\resizebox{0.85\linewidth}{!}{
    \begin{tabular}{l|c|c|c}
        \toprule
        &\multicolumn{3}{c}{PB\_T50\_RS}\\
       Method & PreTr. on MVPNet & PreTr. on MVPNet-\textit{small} & PreTr. on CO3D\\
        \midrule
        PointNet++ & 78.90 / 77.11& \textbf{79.00} / 76.78 & 78.79 / \textbf{77.20}\\
        CurveNet & \textbf{83.68 / 81.17} & 73.92 / 68.92& 73.78 / 69.62\\
        \midrule
        PointMAE & \textbf{84.13 / 81.92} & 81.85 / 79.25  & 81.40 / 78.87  \\ 
        \bottomrule
    \end{tabular}
}
\caption{\textbf{ScanObjectNN real-world point cloud classification} comparison with CO3D \cite{co3d}. Pretrain on MVImgNet, MVImgNet-\textit{small} or CO3D, and finetune on ScanObjectNN. The metric is \textbf{overall / average accuracy}.}
\label{tab:quan_comp_pcd_co3d} 
\vspace{-0.5cm}
\end{table}

\subsection{Self-supervised Point Cloud Pretraining}
\label{sub:pointmae}

\paragraph{Pre-review.}
Self-supervised learning has been exploited for 3D object point cloud understanding \cite{pointbert,OcCo,pointmae,liu2022masked}.
Nevertheless, they are all pretrained on synthetic datasets \cite{modelnet,shapenet}, making it hard to obtain rich real-world representations.
To solve this, our MVPNet becomes a natural choice.

\paragraph{What can MVPNet do?}
In \cref{tab:quan_comp_pcd}, PointMAE pretrained on our MVPNet outperforms the state-of-the-art methods when finetuned on ScanObjectNN, proving the benefit of MVPNet on the self-supervised learning regime for real-world point cloud classification.
\cref{tab:quan_comp_pcd_co3d} also shows that pretraining on MVPNet is more powerful than CO3D \cite{co3d} for real-world point cloud classification.


\subsection{MVPNet Benchmark Challenge}
\label{sec:benchmark}
We present the MVPNet benchmark challenge for real-world point cloud classification, which contains 64,000 training and 16,000 testing samples.
The results of various methods are shown in \cref{tab:mvimgnet_pcd_bench}.

\paragraph{MVPNet is more challenging than ScanObjectNN.}
We firstly train models on ScanObjectNN, then conduct the test on MVPNet, which is concluded in \cref{tab_MVP_challenge}.
On the opposite, we also train on MVPNet and test on ScanObjectNN, which is listed in \cref{tab_MVP_challenge2}.
Comparing \cref{tab_MVP_challenge} and \cref{tab_MVP_challenge2},
the \textit{accuracy drops} are significantly \textit{larger} when training on ScanObjectNN and testing on MVPNet, which verifies the greater challenge of our MVPNet.
All the experiments in 3D understanding strictly follow the original settings of the selected backbone networks.

\begin{table}\centering
    \scriptsize
    \begin{tabular}{c|c}
     \toprule
    Method &  Overall / Average Accuracy\\
    \midrule
    PointNet~\cite{qi2017pointnet} & 70.72 / 54.46\\
    PointNet++~\cite{qi2017pointnet++} & 79.15 / 58.24\\
    DGCNN~\cite{wang2019dynamic} & 86.49 / 63.98\\
    PointMLP~\cite{ma2022rethinking} & 88.89 / 73.64 \\
    CurveNet~\cite{xiang2021walk} & 88.88 / 75.37\\
    GDANet~\cite{xu2021learning} &89.54 / 68.41\\
    PAConv~\cite{xu2021paconv} &83.35 / 59.13\\
    PCT~\cite{guo2021pct} &91.49 / 75.41\\
    \bottomrule
    \end{tabular}
    \caption{\label{tab:mvimgnet_pcd_bench} Quantitative results on our \textbf{new MVPNet benchmark for real-world point cloud classification}.}
    \vspace{-0.2cm}
\end{table}

\begin{table}[t]
    \centering
    \resizebox{0.8\linewidth}{!}{
    \begin{tabular}{c|c|c}
    \toprule
           & train: MVPNet & train: ScanObjectNN\\
    Method & test: MVPNet & test: MVPNet\\
    \midrule
    PointNet++~\cite{qi2017pointnet++} & 79.15 / 58.24 & 15.13 \textcolor{red}{(-64.06)} / 7.96 \textcolor{red}{(-50.28)}\\
    CurveNet~\cite{xiang2021walk} & 88.88 / 75.37 & 46.95 \textcolor{red}{(-41.93)} / 28.24 
 \textcolor{red}{(-47.13)} \\
    \bottomrule
    \end{tabular}
    }
    \caption{Quantitative comparison while \textbf{training on ScanObjectNN, testing on MVPNet}. The evaluation metric is \textbf{overall / average accuracy}.}
    \label{tab_MVP_challenge}
    \vspace{-0.2cm}
\end{table}

\begin{table}[t]
    \centering
    \resizebox{0.8\linewidth}{!}{
    \begin{tabular}{c|c|c}
    \toprule
           & train: ScanObjectNN & train: MVPNet\\
    Method & test: ScanObjectNN & test: ScanObjectNN\\
    \midrule
    PointNet++~\cite{qi2017pointnet++} & 76.50 / 73.42 & 40.35 \textcolor{red}{(-36.15)} / 33.59 \textcolor{red}{(-39.83)}\\
    CurveNet~\cite{xiang2021walk} & 73.96 / 69.96 & 51.84 \textcolor{red}{(-22.12)} / 46.27 \textcolor{red}{(-23.69)}\\
    \bottomrule
    \end{tabular}
    }
    \caption{Quantitative comparison while \textbf{training on MVPNet, testing on ScanObjectNN}. The evaluation metric is \textbf{overall / average accuracy}.}
    \label{tab_MVP_challenge2}
    \vspace{-0.4cm}
\end{table}

\textbf{\textit{MVPNet is suggested for investigating 3D point cloud understanding in the future.}}

\section{Conclusion}%
\label{sec:Conclusion}
We have introduced MVImgNet, a large-scale dataset of multi-view images, which is efficiently collected by shooting videos of real-world objects. The multi-view nature endows our dataset with 3D-aware visual signals, making MVImgNet a soft bridge to link 2D and 3D vision.
To probe the power of MVImgNet, we conduct a host of pilot experiments on various visual tasks, including radiance field reconstruction, multi-view stereo, and view-consistent image understanding, where MVImgNet demonstrates promising effectiveness, expecting more future explorations.
As a bonus of MVImgNet, a point cloud dataset -- MVPNet is derived. 
Experiments show MVPNet can benefit real-world 3D object classification. As a broader impact on society, our datasets \textit{delineates} a world -- that is closer to a colorful and vivid real \textit{3D} world -- where we human lives.

\paragraph{Limitations.}
We mainly focus on human-centric classes, and our data can definitely satisfy the understanding of common objects in human daily life.
As a result, the category richness of MVImgNet is less than ImageNet \cite{imagenet}, which may cause inferior performance when serving for recognition on nature-centric classes such as turtle, bear, \etc.

Moreover, our data do not consider very complex backgrounds, so they can not be \textit{straightforwardly} adopted for complicated scene-level understanding.
Yet we believe that such an issue can be solved by future methods via utilizing some mediate methods such as domain adaptation or knowledge distillation, with the help of \textit{commonsense} knowledge and \textit{universal} representations gained from our data.

\clearpage
\section*{Contributions}%
\label{sec:Contributions}

\noindent\textbf{Xianggang Yu} contributes to the whole building pipeline of the dataset, including data acquisition and data processing. He conducted experiments on radiance field reconstruction. He also designed and advised the experiment on view-consistent SOD.

\noindent\textbf{Mutian Xu} advised the exploration of view-consistent image understanding and 3D understanding. He also organized and wrote the whole paper, and led most part of the research.

\noindent\textbf{Yidan Zhang} worked on data processing. He implemented experiments on view-consistent image classification and view-consistent SOD. He also collected and provided illustrations of dataset statistics and dataset details. 

\noindent\textbf{Haolin Liu} was responsible for point cloud labeling and cleaning. He implemented experiments on 3D point cloud classification.

\noindent\textbf{Chongjie Ye} conducted data labelling. He implemented experiments on multi-view stereo and 3D point cloud classification.

\noindent\textbf{Yushuang Wu} worked on the data collection and annotation. He also collected and wrote related works.

\noindent\textbf{Zizheng Yan} suggested the experiments on view-consistent image understanding.

\noindent\textbf{Chenming Zhu} conducted the experiment on view-consistent contrastive learning.

\noindent\textbf{Zhangyang Xiong} worked on data collection and processing.

\noindent\textbf{Tianyou Liang} worked on data cleaning and assisted experiments on view-consistent image understanding.

\noindent\textbf{Guanying Chen, Shuguang Cui} advised the project.

\noindent\textbf{Xiaoguang Han} proposed and led the whole research.

\clearpage
{\small

\bibliographystyle{ieee_fullname}
}

\clearpage
\setcounter{figure}{0}
\setcounter{table}{0}

\begin{appendices}
\startcontents[supple]

{
    \hypersetup{linkcolor=black}
    \printcontents[supple]{}{1}{}
}
\renewcommand{\thefootnote}{\fnsymbol{footnote}}
\renewcommand{\thesection}{\Alph{section}}%
\renewcommand\thetable{\Roman{table}}
\renewcommand\thefigure{\Roman{figure}}

\newcommand{\todo}[1]{\textcolor{red}{{[TODO: #1]}}}




\section{Per-category Data Distribution}
\begin{figure}[t]
    \centering
    \includegraphics[width=\columnwidth]{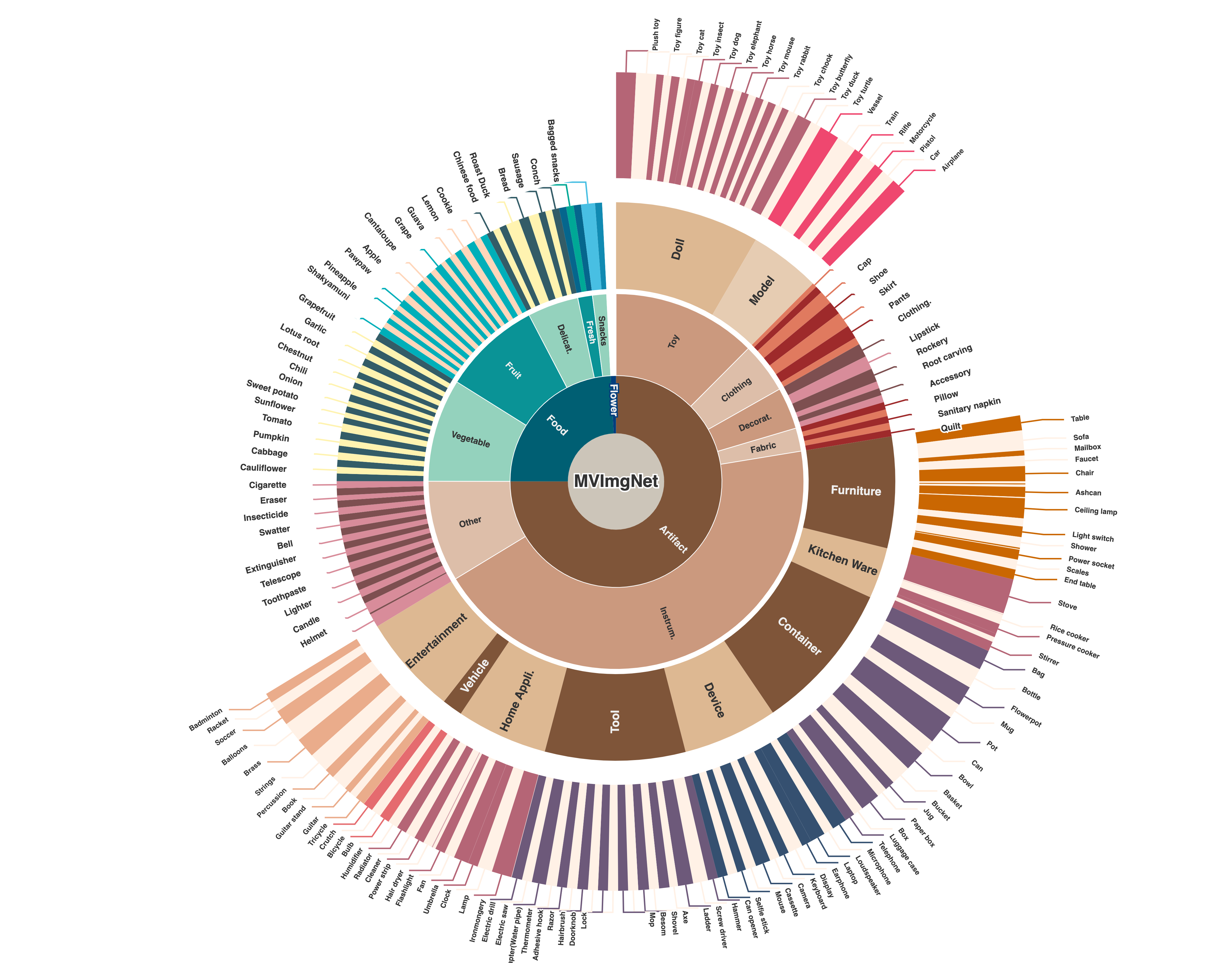}
    \caption{\textbf{Category taxonomy of MVImgNet}, where the angle of each class denotes its actual data proportion. \textbf{Interior}: Parent class. \textbf{Exterior}: Children class.}
    \label{fig:taxonomy}
\end{figure}

\begin{figure}[t]
  \centering
    \includegraphics[width=\columnwidth]{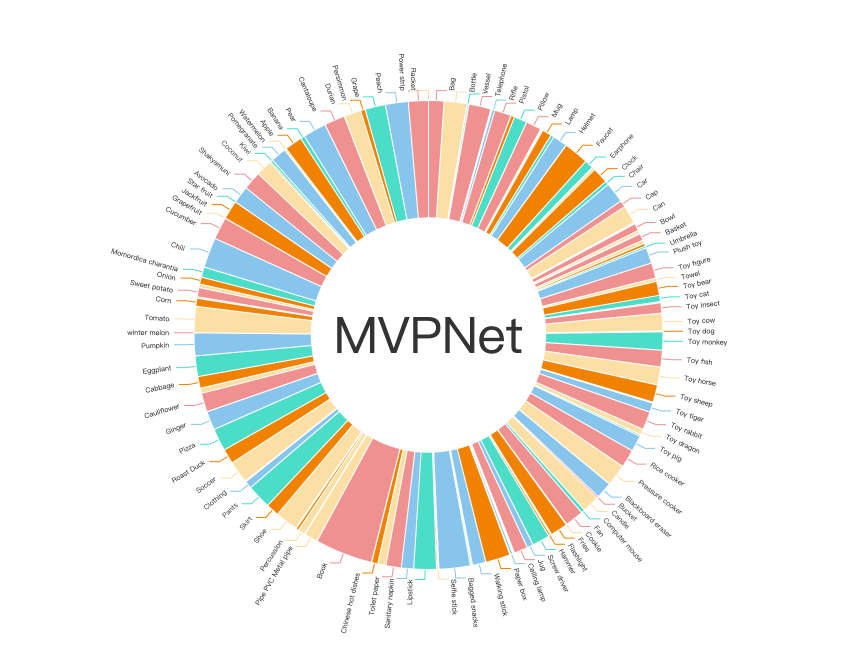}
    \caption{\label{fig:MVP_taxonomy}\textbf{Category distribution of MVPNet.}}
\end{figure}

\begin{figure*}[t]
  \centering
   \includegraphics[width=0.82\linewidth]{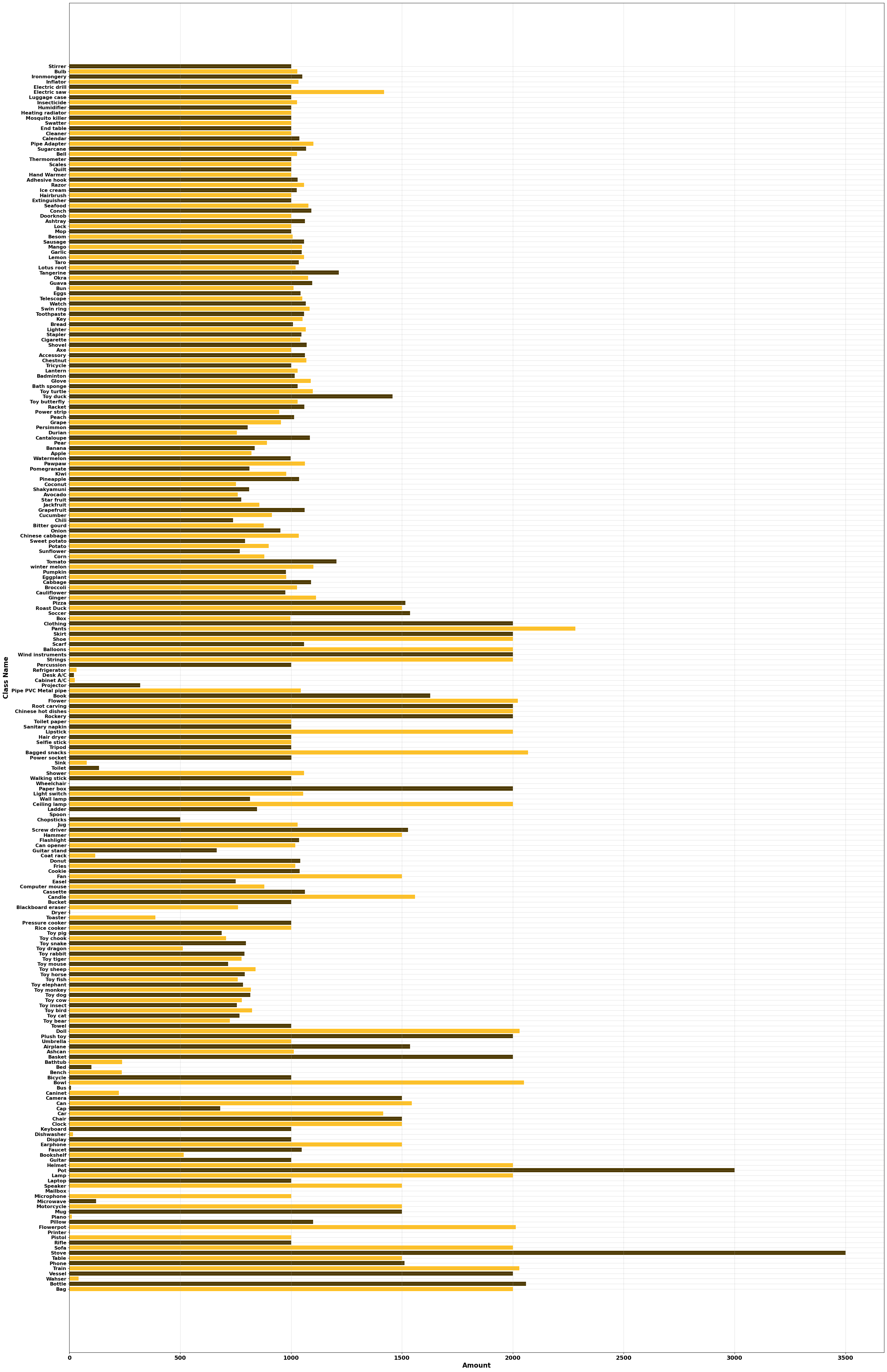}
   \caption{Data amount of each category of \textbf{MVImgNet}.} 
   \label{fig:sup_mvi_ds}
\end{figure*}

\begin{figure*}[t]
  \centering
   \includegraphics[width=0.82\linewidth]{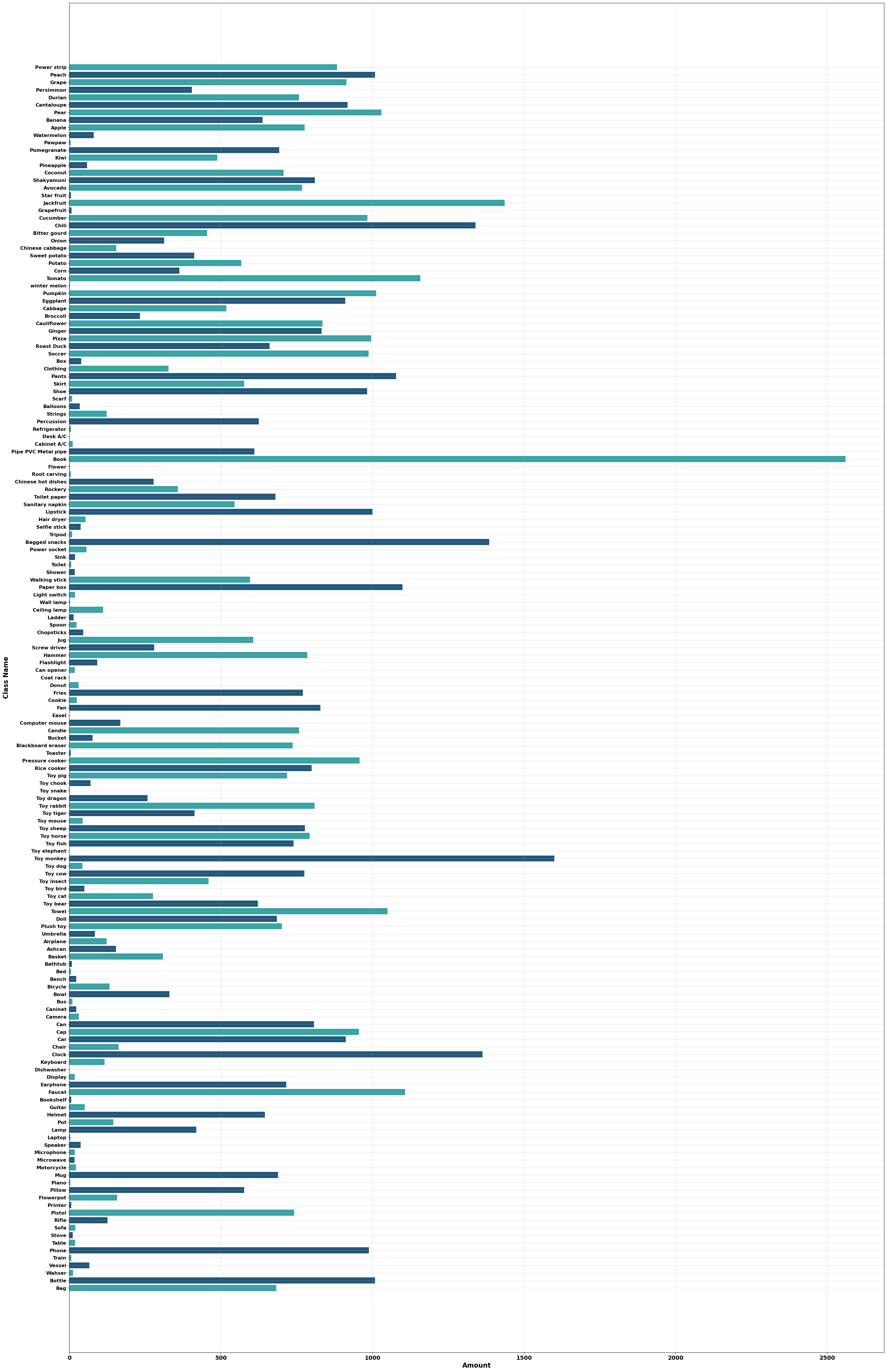}
   \caption{Data amount of each category in \textbf{MVPNet}.} 
   \label{fig:sup_mvp_ds}
\end{figure*}

The category taxonomy is shown in Fig.~\ref{fig:taxonomy} for MVImgNet, and  Fig.~\ref{fig:MVP_taxonomy} for MVPNet.
The per-category data distribution is illustrated in Fig.~\ref{fig:sup_mvi_ds} for MVImgNet, and Fig.~\ref{fig:sup_mvp_ds} for MVPNet. 
The average size is 921 per class for MVImgNet and 581 per class for MVPNet. 

\section{More Visualizations of Data Samples}

\begin{figure*}[t]
  \centering
   \includegraphics[width=0.85\linewidth]{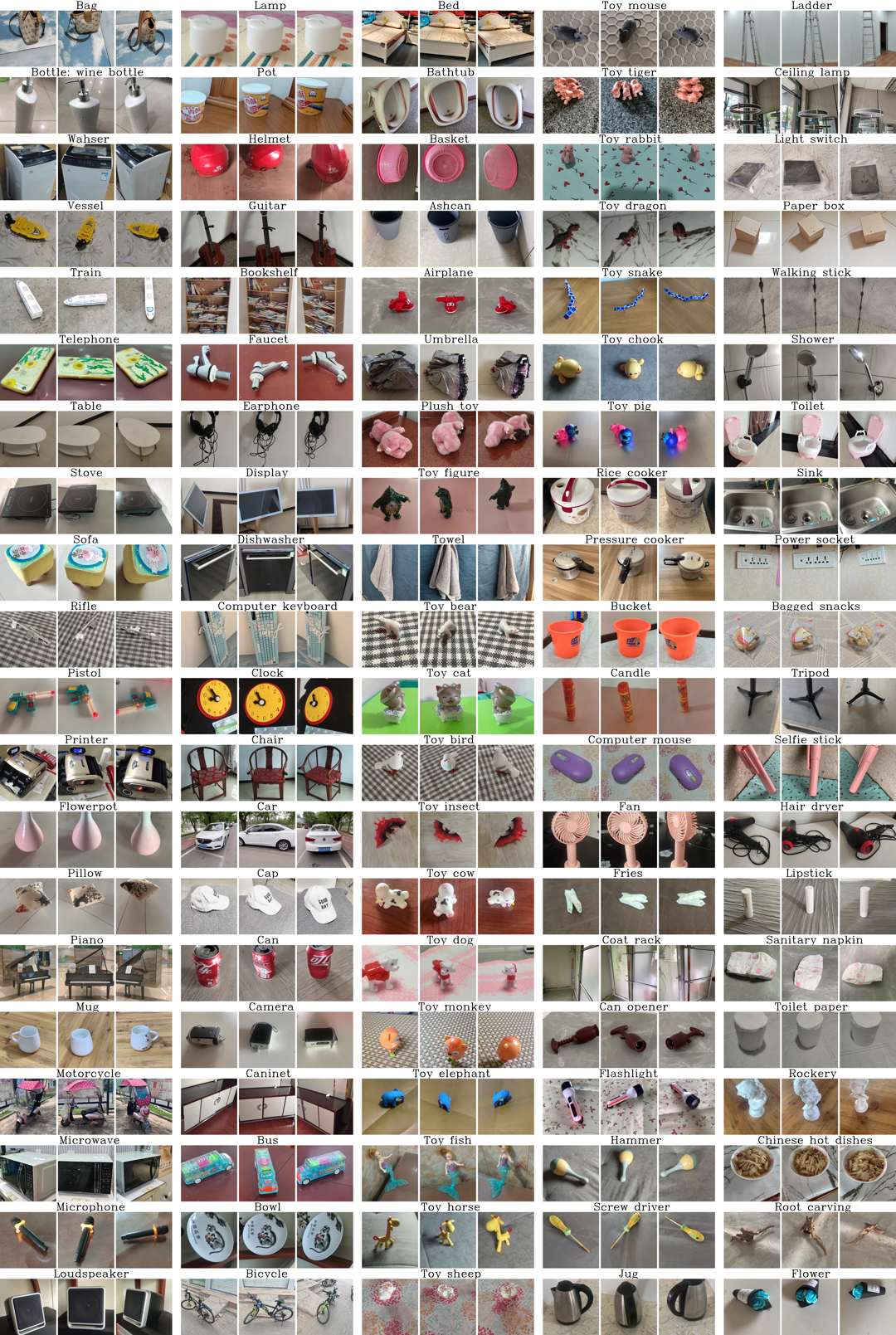}
   \caption{A variety of multi-view images in \textbf{MVImgNet}.} 
   \label{fig:sup_mvi_da}
\end{figure*}

\paragraph{MVImgNet.}
Fig.~\ref{fig:sup_mvi_da} presents a larger set of examples in MVImgNet. 
Several multi-view images and the corresponding class label are illustrated for each sample.
It clearly shows the differences between each view, and comprehensive categories in our dataset.

\begin{figure*}[t]
  \centering
   \includegraphics[width=0.85\linewidth]{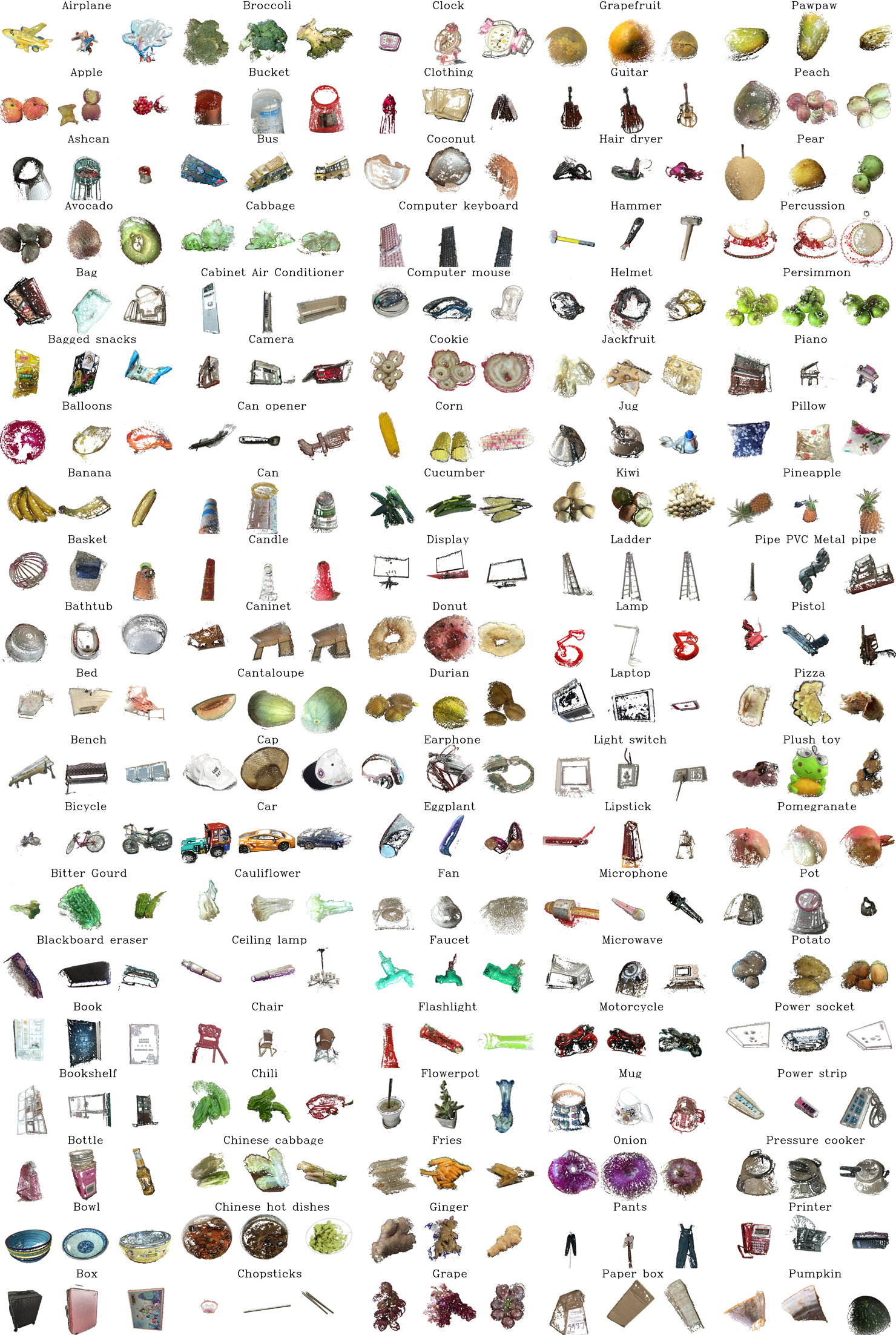}
   \caption{A variety of 3D object point clouds in \textbf{MVPNet}.} 
   \label{fig:sup_mvp_da}
\end{figure*}

\paragraph{MVPNet.}
Fig.~\ref{fig:sup_mvp_da} shows various 3D point clouds from MVPNet.
It can be seen that each sample has a distinct texture, noise, and pose, indicating real-world signals.

\section{More Visualizations of Qualitative Results}

\paragraph{Radiance field reconstruction.}
We visualize more results of generalizable NeRF reconstruction in Fig.~\ref{fig:nerf_suppl}, where the MVImgNet-pretrained model performs consistently much better than the train-from-scratch model.
\newcommand{\supplgtfigwidth}{0.1\textwidth}
\newcommand{\supplresultswidth}{0.125\textwidth}

\newcommand{\supplinsertimg}[1]{
  \makecell{
  \includegraphics[width=\supplresultswidth]{#1} \\
  }
}

\begin{figure*}[t]
	\centering
	\scriptsize
	\begin{tabular}{@{}c@{\,\,}c@{}c@{}c@{}@{}c@{\,\,}c@{}c@{}c@{}}
		\makecell[c]{
			\includegraphics[trim={0px 0px 0px 0px}, clip, width=\supplgtfigwidth]{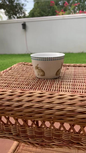}
			\\
			\textit{Bowl}
		}
		&
		\supplinsertimg{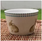} &
		\supplinsertimg{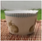} &
		\supplinsertimg{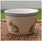} &
		\makecell[c]{
			\includegraphics[trim={0px 0px 0px 0px}, clip, width=\supplgtfigwidth]{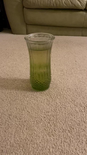}
			\\
			\textit{Vase}
		}
		&
		\supplinsertimg{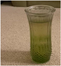} &
		\supplinsertimg{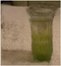} &
		\supplinsertimg{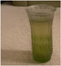} \\
		\makecell[c]{
			\includegraphics[trim={0px 0px 0px 0px}, clip, width=\supplgtfigwidth]{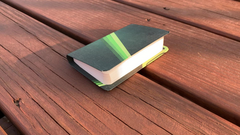}
			\\
			\textit{Book}
		}
		& 
		\supplinsertimg{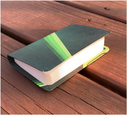} &
		\supplinsertimg{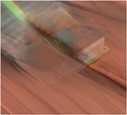} &
		\supplinsertimg{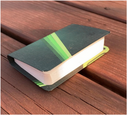} &
		\makecell[c]{
			\includegraphics[trim={0px 0px 0px 0px}, clip, width=\supplgtfigwidth]{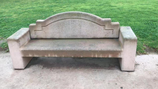}
			\\
			\textit{Bench}
		}
		& 
		\supplinsertimg{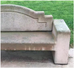} &
		\supplinsertimg{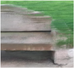} &
		\supplinsertimg{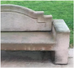} \\
		\makecell[c]{
			\includegraphics[trim={0px 0px 0px 0px}, clip, width=\supplgtfigwidth]{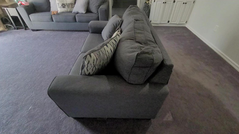}
			\\
			\textit{Couch}
		}
		& 
		\supplinsertimg{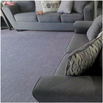} &
		\supplinsertimg{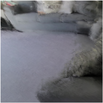} &
		\supplinsertimg{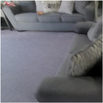} &
		\makecell[c]{
			\includegraphics[trim={0px 0px 0px 0px}, clip, width=\supplgtfigwidth]{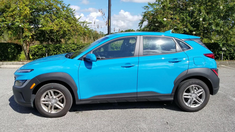}
			\\
			\textit{Car}
		}
		&
		\supplinsertimg{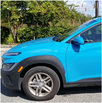} &
		\supplinsertimg{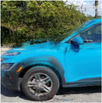} &
		\supplinsertimg{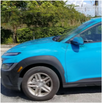} \\
		\makecell[c]{
			\includegraphics[trim={0px 0px 0px 0px}, clip, width=\supplgtfigwidth]{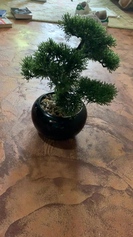}
			\\
			\textit{Plant}
		}
		& 
		\supplinsertimg{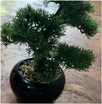} &
		\supplinsertimg{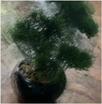} &
		\supplinsertimg{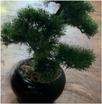} &
		\makecell[c]{
			\includegraphics[trim={0px 0px 0px 0px}, clip, width=\supplgtfigwidth]{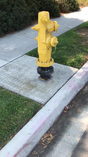}
			\\
			\textit{Hydrant}
		}
		&
		\supplinsertimg{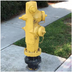} &
		\supplinsertimg{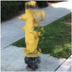} &
		\supplinsertimg{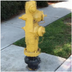} \\
		\makecell[c]{
			\includegraphics[trim={0px 0px 0px 0px}, clip, width=\supplgtfigwidth]{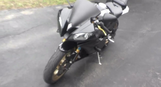}
			\\
			\textit{Motorcycle}
		}
		& 
		\supplinsertimg{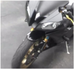} &
		\supplinsertimg{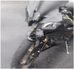} &
		\supplinsertimg{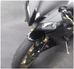} &
		\makecell[c]{
			\includegraphics[trim={0px 0px 0px 0px}, clip, width=\supplgtfigwidth]{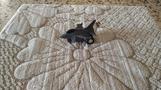}
			\\
			\textit{Toyplane}
		}
		& 
		\supplinsertimg{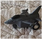} &
		\supplinsertimg{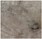} &
		\supplinsertimg{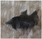} \\
		\makecell[c]{
			\includegraphics[trim={0px 0px 0px 0px}, clip, width=\supplgtfigwidth]{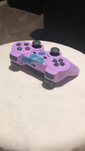}
			\\
			\textit{Remote}
		}
		& 
		\supplinsertimg{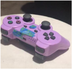} &
		\supplinsertimg{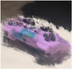} &
		\supplinsertimg{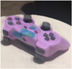} &
		\makecell[c]{
			\includegraphics[trim={0px 0px 0px 0px}, clip, width=\supplgtfigwidth]{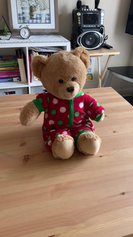}
			\\
			\textit{Teddybear}
		}
		&
		\supplinsertimg{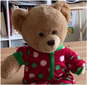} &
		\supplinsertimg{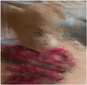} &
		\supplinsertimg{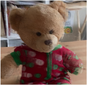} \\
		\makecell[c]{
			\includegraphics[trim={0px 0px 0px 0px}, clip, width=\supplgtfigwidth]{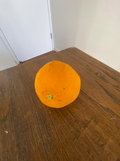}
			\\
			\textit{Orange}
		}
		&
		\supplinsertimg{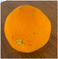} &
		\supplinsertimg{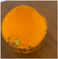} &
		\supplinsertimg{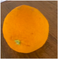} &
		\makecell[c]{
			\includegraphics[trim={0px 0px 0px 0px}, clip, width=\supplgtfigwidth]{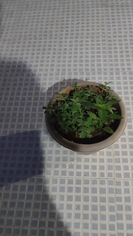}
			\\
			\textit{Plant}
		}
		&
		\supplinsertimg{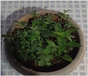} &
		\supplinsertimg{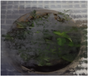} &
		\supplinsertimg{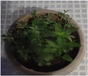} \\
		& Ground Truth  &  Scratch & \textbf{MVImgNet-pretrained} & & Ground Truth & Scratch & \textbf{MVImgNet-pretrained}
	\end{tabular}
	\caption{More qualitative comparison on real-world 360$^\circ$ objects~\cite{co3d} of \textbf{MVImgNet-pretrained} IBRNet~\cite{wang2021ibrnet} model and the \textbf{train-from-scratch} model.}
	
	\label{fig:nerf_suppl}
\end{figure*}



\begin{figure*}[t]
  \centering
   \includegraphics[width=1\textwidth]{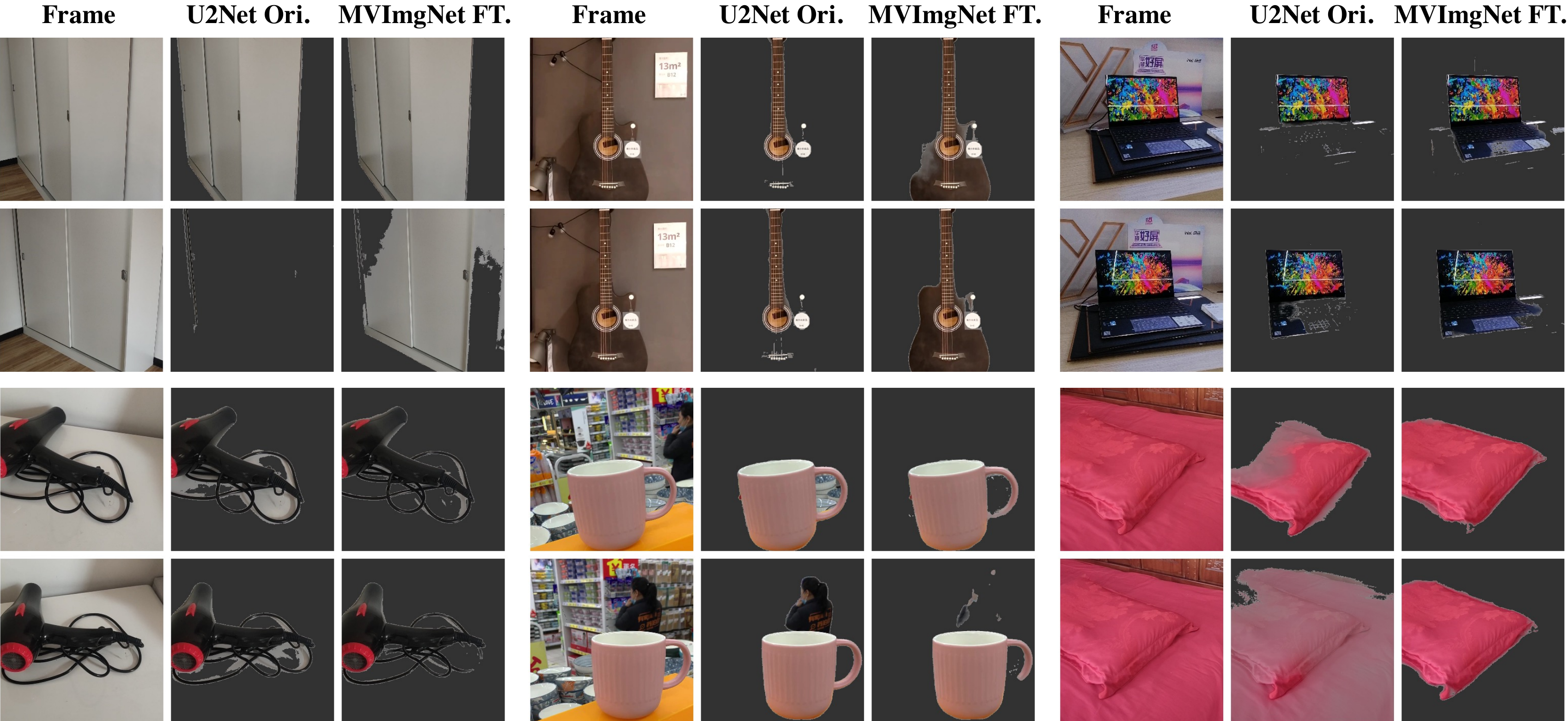}
   \caption{More qualitative results of view-consistent salient object detection. \textbf{Finetuning U2Net \cite{Qin_2020_PR} on MVImgNet} improves the performance.}
   \label{fig:sup_sod_res}
\end{figure*}

\paragraph{View-consistent SOD.}
Fig.~\ref{fig:sup_sod_res} illustrates more results of the view-consistent salient objection detection (SOD) task on our MVImgNet test set, where finetuning U2Net \cite{Qin_2020_PR} on MVImgNet gains better result than the original U2Net.

\section{More Experiments of Data Scalability}

As indicated in the main paper, more power can be gained with more data utilized from our datasets. In this section, we provide more experimental results following such rules.

\paragraph{Multi-view stereo.}
Tab.~\ref{tab:supple_mvs} lists the MVS depth map accuracy on DTU \cite{dtu} evaluation set.
It shows that using larger amounts of videos from MVImgNet for pretraining yields higher accuracy.

\paragraph{View-consistent image classification.}
Similar conclusions are also found in the view-consistent image classification task.
We progressively add more MVImgNet training data into MVI-Mix data (mixing the original ImageNet \cite{imagenet} data with MVImgNet data as stated in the main paper) to train ResNet-50 \cite{he2016deep} and evaluate on MVImgNet test set. 
Tab.~\ref{tab:supple_img_cls} demonstrates that adding more MVImgNet training data brings better view consistency for the image recognition task.

\begin{table}[t]
	\centering
        \scriptsize
	\begin{tabular}{l|ccc}
\toprule
Method & $2mm\uparrow$ & $4mm\uparrow$ & $8mm\uparrow$  \\ 
\midrule
pretrained with 10k videos & 52.96  & 72.25 & 83.79   \\ 
pretrained with 50k videos &  56.86 & 73.79 & 83.42\\ 
pretrained with 100k videos &   58.63 & 75.20 &   84.28 \\   \bottomrule
\end{tabular}
	\caption{\textbf{MVS depth map accuracy} on DTU \cite{dtu} evaluation set, using \textbf{different amounts (10k, 50k, 100k) of videos} (one video may contain several multi-view images / frames) from MVImgNet for pretraining.}
	\label{tab:supple_mvs} 
\end{table}

\begin{table}[t]
  \centering
  \scriptsize
  \begin{tabular}{cccc}
    \toprule
    Scale  & Confidence Var & Accuracy \\
    \midrule
    ImageNet-only                & 0.207  & 53.09\% \\ 
    MVI-Mix with 20k videos      & 0.119  & 75.03\% \\
    MVI-Mix with 40k videos      & 0.114  & 76.88\% \\
    MVI-Mix with 80k videos      & 0.104  & 77.03\% \\
    MVI-Mix with 100k videos     & 0.102  & 77.31\%\\
    MVI-Mix with 120k videos     & 0.101  & 77.47\%\\
    \bottomrule
  \end{tabular}
  \caption{\textbf{View-consistency image classification results} on MVImgNet test set, using \textbf{different amounts (20k, 40k, 80k, 100k, 120k) of videos} (one video may contain several multi-view images / frames) from MVImgNet for training ResNet-50 \cite{he2016deep} (smaller Confidence Var and higher Accuracy indicate better view consistency).}
  \label{tab:supple_img_cls}
\end{table}

\paragraph{Real-world point cloud classification.}
Besides, as shown in Tab.~\ref{tab:scalability_pc_cls_1} and Tab.~\ref{tab:scalability_pc_cls_2}, when employing larger ratios of data from MVPNet for pretraining both supervised (\ie, PointNet++ \cite{qi2017pointnet++}, CurveNet \cite{xiang2021walk}) and self-supervised models (\ie, PointMAE \cite{pointmae}), the better performance can be achieved when fine-tuning them on ScanObjectNN dataset \cite{scanobjectnn} for real-world point cloud classification task.

\section{More Discussions about Our Datasets}

\paragraph{Data filter.}
Our $\sim$219k videos are screened from $\sim$260k raw videos, where the videos with bad camera estimations are filtered. 
When building MVPNet, we select 90k (the most common 150 categories are chosen) videos, yielding 87k point clouds to remain after the manual cleaning.

\paragraph{Real-world captures.}
Note that when we capture the object videos, we maintain the \textit{original} status of objects in \textit{real-world} environments, \ie, objects will \textit{not be intentionally} displayed standalone for ideal $360^\circ$ captures (\eg, the sofa is against the wall).  
By doing so:
{\bf 1)}~The capture is easy to conduct, making it possible to build a very large-scale dataset.
{\bf 2)}~The produced data better matches the \textit{real-world applications}, \eg, our obtained point clouds are usually of partial views which are more like real-captured. 
{\bf 3)}~The produced images usually contain the diverse scene-level \textit{background}, instead of the $360^\circ$ capture of single objects on a \textit{clean} supporter.
This better provides the potential for \textit{in-the-wild} scene-level visual tasks.

\begin{table}[t]
    \centering
    \resizebox{0.99\linewidth}{!}{
    \begin{tabular}{c|c|c|c|c}
    \toprule
    & \multicolumn{4}{c}{Add Random Rotation}\\
    Method & from scratch& \textbf{25\%}& \textbf{50\%} & \textbf{100\%}\\
    \midrule
    PointNet++~\cite{qi2017pointnet++} & 76.50 / 73.42&77.82 / 75.98 & 78.11 / 76.13  &78.76/76.54\\
    CurveNet~\cite{xiang2021walk} & 73.96 / 69.96 & 73.75 / 69.86 &75.83 / 72.48 &78.99 / 76.59\\
    \midrule
    PointMAE ~\cite{pointmae} & 83.17 / 80.75 & 83.83 / 81.94 & 85.22 / 83.34 & 86.19 / 84.60\\ 
    \bottomrule
    \end{tabular}
    }
    \caption{\textbf{ScanObjectNN \cite{scanobjectnn} real-world point cloud classification results} of using \textbf{different ratio (25\%, 50\%, 100\%) of data from MVPNet for pretraining} under the setting of Add Random Rotation. The metric is \textbf{overall / average accuracy}.}
    \label{tab:scalability_pc_cls_1}
\end{table}

\begin{table}[t]
    \centering
    \resizebox{0.99\linewidth}{!}{
    \begin{tabular}{c|c|c|c|c}
    \toprule
    &\multicolumn{4}{c}{PB\_T50\_RS}\\
    Method & from scratch& \textbf{25\%}& \textbf{50\%} & \textbf{100\%}\\
    \midrule
    PointNet++~\cite{qi2017pointnet++}&78.80 / 75.70&79.67 / 76.63&81.36 / 79.33&80.22 / 76.91\\
    CurveNet~\cite{xiang2021walk}&74.27 / 69.43&77.26 / 72.65&81.32 / 78.03&83.68 / 81.17\\
    \midrule
    PointMAE ~\cite{pointmae}& 77.34 / 73.52 & 82.75 / 79.90 & 84.18 / 81.41 & 84.13 / 81.92 \\
    \bottomrule
    \end{tabular}
    }
    \caption{\textbf{ScanObjectNN \cite{scanobjectnn} real-world point cloud classification results} of using \textbf{different ratio (25\%, 50\%, 100\%) of data from MVPNet for pretraining} under the setting of PB\_T50\_RS. The metric is \textbf{overall / average accuracy}.)}
    \label{tab:scalability_pc_cls_2}
\end{table}

\section{Implementation Details}

\subsection{3D Reconstruction}

\paragraph{Radiance field reconstruction.}
We choose IBRNet~\cite{wang2021ibrnet} as the baseline method, and use the original training datasets of IBRNet \cite{wang2021ibrnet}, 
which include Google Scanned Objects~\cite{gso}, RealEstate10K~\cite{RealEstate10K}, the Spaces dataset~\cite{flynn2019deepview}, and 102 real scenes from handheld cellphone captures. 
We pretrain IBRNet on the full MVImgNet dataset and finetune on the aforementioned IBRNet training datasets for 10k iterations. For each object, 8$\sim$12 views are used for training and 10 views for inference. \#views is independent on \#objects.
The raw input resolution of each sample is used for computing, and it varies.
The finetuning takes 10k iterations, and the scratch model is exactly the same as the author-released IBRNet model for a fair comparison.
The pretraining takes about 3 days on 8 RTX3090 GPUs.

\paragraph{Multi-view stereo.}
Multi-view stereo (MVS) aims at recovering 3D scenes from multi-view images and calibrated cameras.
As for the data preprocessing, 200K frames are randomly sampled from 100K videos in MVImgNet, and are resized to 640 $\times$ 360 or 360 $\times$ 640.
We choose JDACS \cite{xu2021self} to perform self-supervised pretraining on MVImgNet. 
JDACS takes multi-view images and corresponding poses as input, and uses MVSNet as the backbone to output the synthetic/pseudo depth, where the self-supervision signal is provided by multi-view consistency.
%

\subsection{View-consistent Image Understanding}

\paragraph{View-consistent image classification.}

As mentioned in the main paper, we mix MVImgNet and original ImageNet \cite{imagenet} for creating a new training set.
The hybrid datasets contain 1, 100 categories (after removing the overlapping classes), coming from 500k frames of 100k MVImgNet videos and 200k ImageNet images.

\paragraph{View-consistent contrastive learning.}
We follow the original MoCo v2 to conduct experiments. 
For reducing view redundancy, we randomly sample 5 frames of each video from MVImgNet for finetuning. 
For each iteration, we randomly sample two view images from the same video as positive pair and apply random data augmentation to increase the generalization capability of the model, images from other videos will be treated as negative pairs

\textbf{View-consistent SOD.}
We propose to leverage the multi-view consistency to improve SOD with the help of \textit{optical flows}.
The two adjacent frames should be the same after warping the optical flow to one of the other frames, 
yielding the loss of the optical flow as:

\begin{equation}
        Loss_{OF} = \mathcal{M}(f_i) - \mathcal{M}(f_{i-1}) \cdot \mathcal{F}(f_i),
  \label{eq:loss_maskflow}
\end{equation}
where $i$ denotes the frame index, $\mathcal{M}$ means the mask, and $\mathcal{F}$ is the optical flow between $f_i$ and $f_{i-1}$ calculated before training.
By adding $Loss_{of}$ into the original SOD loss, the final loss is:
\begin{equation}
        Loss = \tau * Loss_{OF} + (1 - \tau) * Loss_{SOD},
  \label{eq:loss_all}
\end{equation}
where $\tau$ is set to 0.15 in our experiments.

For fast training, we sample 10 frames uniformly from each video of 100k MVImgNet and 10, 553 training images from DUTS-TR \cite{dso}.

\subsection{3D Understanding}
All the experiments in 3D understanding strictly follow the original settings of the selected backbone networks.


\stopcontents[supple]

\end{appendices}

\end{document}